%% file: acl_latex.tex
\pdfoutput=1

\documentclass[11pt]{article}

\usepackage[preprint]{acl}

\usepackage{times}
\usepackage{latexsym}

\usepackage[T1]{fontenc}

\usepackage[utf8]{inputenc}

\usepackage{microtype}

\usepackage{inconsolata}

\usepackage{graphicx}

\usepackage{graphicx}
\usepackage{algorithm}
\usepackage{algpseudocode}
\usepackage{booktabs,chemformula}
\usepackage{amsfonts}
\usepackage[subrefformat=parens]{subcaption}
\usepackage{adjustbox}
\usepackage{cleveref}
\usepackage{makecell}
\usepackage{xcolor}
\usepackage{colortbl}
\usepackage{multirow}
\usepackage{mathabx}
\definecolor{maroon}{cmyk}{0.60, 0.19, 0.0, 0.08}
\definecolor{redd}{cmyk}{0.19, 0.60, 0.0, 0.08}
\usepackage{xcolor, tabularx, booktabs}

\usepackage{tcolorbox}

\title{On Efficient Language and Vision Assistants for Visually-Situated Natural Language Understanding: What Matters in Reading and Reasoning}

\author{Geewook Kim \\
  NAVER Cloud AI\\
  KAIST AI\\
  \texttt{gwkim.rsrch@gmail.com} \\\And
  Minjoon Seo \\
  KAIST AI\\
  \texttt{minjoon@kaist.ac.kr} \\}

\begin{document}
\maketitle
\begin{abstract}
Recent advancements in language and vision assistants have showcased impressive capabilities but suffer from a lack of transparency, limiting broader research and reproducibility. While open-source models handle general image tasks effectively, they face challenges with the high computational demands of complex visually-situated text understanding. Such tasks often require increased token inputs and large vision modules to harness high-resolution information. Striking a balance between model size and data importance remains an open question. This study aims to redefine the design of vision-language models by identifying key components and creating efficient models with constrained inference costs. By strategically formulating datasets, optimizing vision modules, and enhancing supervision techniques, we achieve significant improvements in inference throughput while maintaining high performance. Extensive experiments across models ranging from 160M to 13B parameters offer insights into model optimization.
We will fully open-source our codebase, models, and datasets at \url{https://github.com/naver-ai/elva}.
\end{abstract}

\section{Introduction}

\begin{figure}[!t]
    \centering
    \includegraphics[width=\linewidth]{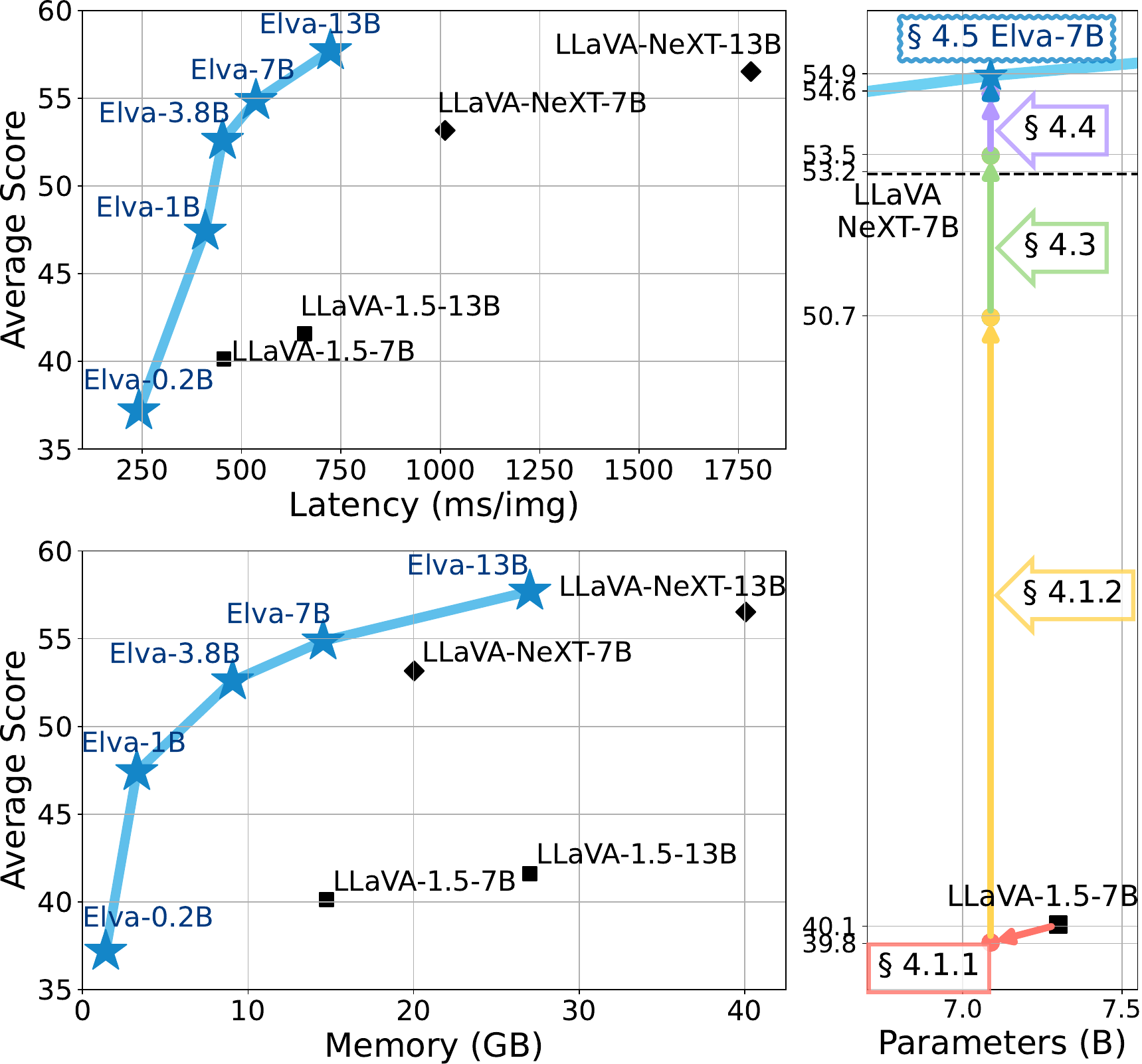}
\caption{\textbf{Graphical comparison illustrating average score against latency and memory consumption for various models.} Scores are derived from eight benchmarks: DocVQA~\citep{mathew2021docvqa}, ChartQA~\citep{masry-etal-2022-chartqa}, InfographicVQA~\citep{Mathew_2022_WACV}, SEED-IMG~\citep{li2023seed}, SEED-2-Plus~\citep{li2024seed2plus}, MMStar~\citep{chen2024we}, ScienceQA~\citep{lu2022learn}, and HallusionBench~\citep{Guan_2024_CVPR}. See Section \ref{sec:problem_and_improvement} for benchmark details. \textbf{\textsc{Elva}} excels with high performance, reduced latency, and lower memory usage. Right: Performance improvements from LLaVA to \textsc{Elva} at the 7B scale, achieved through strategies in Section \ref{sec:remedies}.}
    \label{fig:teaser}
\end{figure}

Recent advancements in integrating Large Language Models (LLMs) with computer vision have led to the creation of sophisticated Language and Vision Assistants~\citep{liu2023llava,liu2024llavanext,kim-etal-2023-visually,laurencon2024matters,laurençon2024building}. These systems are capable of interpreting text within images, enabling them to excel in complex tasks requiring both visual and textual understanding. Notably, models like GPT-4(V) ~\citep{openai2023gpt4} demonstrate a range of sophisticated capabilities, including visually-situated Natural Language Understanding (NLU) tasks, positioning them as powerful assistants. However, these models also face significant challenges related to transparency and accessibility, limiting broader utilization.

Open-source alternatives such as LLaVA~\citep{liu2023llava,liu2024llavanext} have emerged to address these issues. However, as these models grow in complexity, concerns about their reproducibility and resource efficiency persist. Some open-source models provide only the model weights without comprehensive specifications, making replication and use more challenging.

In the fast-evolving realm of Vision-Language Models (VLMs), simply expanding model size and consuming more resources does not necessarily enhance practical utility. It is crucial to strike a balance between high performance and resource efficiency to democratize access to advanced VLMs. Particularly, inference costs are a significant concern for practitioners developing real-world applications. Despite the importance of this balance, key elements contributing to VLM success are still not fully explored.

Traditionally, to enhance the performance, many VLMs have increased their model resolution, often leading to larger and more resource-intensive models. In this work, we challenge this approach by introducing \textsc{Elva} (\textbf{E}fficient \textbf{L}anguage and \textbf{V}ision \textbf{A}ssistant), a suite of VLMs designed to maintain high performance while reducing inference costs. While we do increase training costs to a manageable extent, the primary research target of \textsc{Elva} is to create models capable of handling high-resolution tasks with low inference costs. 

Our key contributions are as follows:
\begin{enumerate}
    \item \textbf{Efficiency and Reproducibility:} We present \textsc{Elva}, an efficient and scalable model architecture trained on open-source data, demonstrating superior reproducibility and cost-effectiveness as shown in Figure~\ref{fig:teaser}.
    \item \textbf{Empirical Validation:} We conduct thorough experiments to validate the effectiveness of \textsc{Elva}'s primary components.
    \item \textbf{Model Scalability:} We develop \textsc{Elva} versions ranging from 160M to 13B parameters, showcasing its scalability and adaptability.
    \item \textbf{Dataset Contributions:} To test \textsc{Elva} as a document assistant, we introduce two new datasets, CORD-Instruct and Parsing-Bench.
    \item \textbf{Open-Source Initiative:} To foster further community research and ensure model reproducibility, we will fully open-source the trained models and datasets from this study.
\end{enumerate}

Our ultimate goal is to shed light on the complexities of VLMs, helping readers identify the critical factors driving model success while presenting a practical, cost-effective solution for diverse real-world applications. Following this introduction, \S2 provides an overview of the foundational LLaVA framework; \S3 discusses computational challenges; \S4 outlines our proposals; \S5 presents our empirical results and analysis; \S6 offers further analysis and ablations; and \S7, along with \S8, surveys related work and concludes the study, respectively.

\section{Large Language and Vision Assistants}

\begin{figure}[t]
    \centering
    \includegraphics[width=\linewidth]{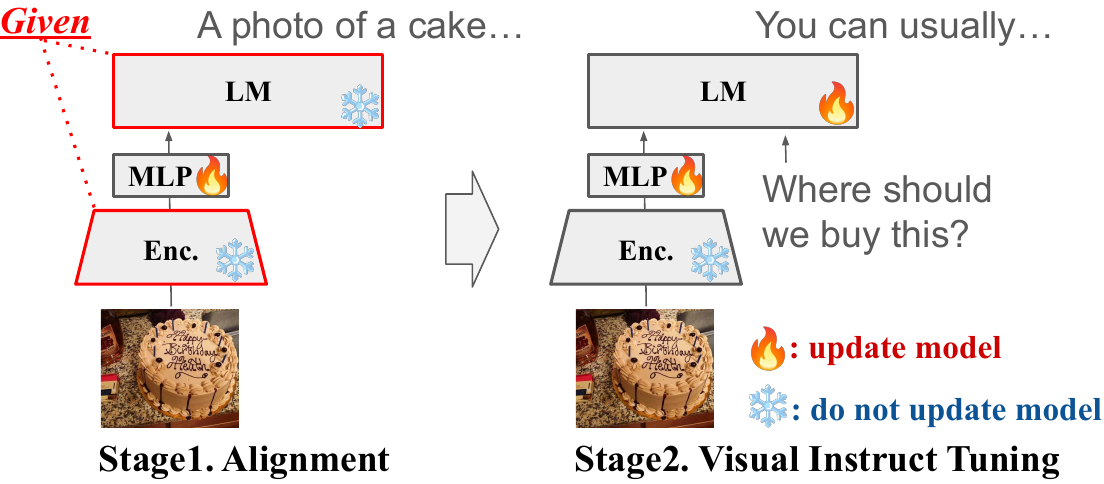}
    \caption{\textbf{Training pipeline consists of two stages.} Alignment of visual and textual features through the MLP, followed by joint training of the LM and the MLP.}
    \label{fig:architecture}
\end{figure}

\paragraph{Architecture.} The LLaVA framework (Figure~\ref{fig:architecture}) employs a pre-trained Vision Transformer (ViT)~\citep{dosovitskiy2020vit} as its vision encoder. Input images are resized and divided into patches of size $n \times (p_h \times p_w \times c)$, where $n = (h/p_h) \times (w/p_w)$, with $h$ and $w$ representing the resized image size, and $p_h$ and $p_w$ denoting the patch size, which are hyperparameters. Here, $c = 3$ denotes the number of channels, typically for RGB images. These patches are processed by the encoder to generate embeddings $\{\mathbf{z}_i \in \mathbb{R}^d\}$, where $d$ is the embedding dimension of the encoder and also a hyperparameter. These embeddings are then mapped to the input space of the language model via a Multi-Layer Perceptron (MLP) before being fed into the model. Optionally, the AnyRes mechanism~\citep{liu2024llavanext} can be applied, allowing for processing of larger images by first segmenting images into $m$ parts to better capture local features. Simultaneously, the entire image is processed to extract global features. Both global and local features are sequentially fed into the language model, allowing it to utilize comprehensive image information. Therefore, the total token count becomes $(1+m) \times n$. See \citet{liu2024llavanext} for more details. However, this approach poses a computational challenge due to the increased token count.

\paragraph{Training Objectives and Datasets.} The model is trained to minimize Cross-Entropy (CE) loss. During pre-training (alignment phase), it generates captions for images, with CE loss computed on the text. In the instruct tuning stage, given an image, question, and answer, the loss is computed on the answer text. The LLaVA-1.5 dataset~\citep{liu2023improvedllava} is widely used and this study aims to further enhance the dataset. More details are in Section~\ref{sec:remedies}.

\section{Efficiency Challenges in LLaVA Models}

This section addresses common overhead issues in LLaVA models, identifying critical limitations and defining the problem space for future work~\citep{liu2024llavanext,internlmxcomposer2_4khd}.

\subsection{Inference Overhead Sources}

Inference overhead in LLaVA models stems from several factors:

\begin{itemize}
    \item \textbf{Model Scale:} Larger models (e.g., 34B parameters) offer enhanced capabilities but incur significant computational costs.
    \item \textbf{Vision Encoder Complexity:} Advanced image encoders like SO400M and ViT-G (1.8B) improve performance but increase overhead~\citep{sun2024evaclip18b,Zhai_2023_ICCV}.
    \item \textbf{Image Resolution:} High resolutions (e.g., 4K) for detailed visual tasks like \textit{DocVQA}~\citep{mathew2021docvqa} increase computational demands on the vision encoder.
    \item \textbf{Vision Token Quantity:} Higher resolutions lead to more vision tokens, increasing the computational load on the LLM (e.g., LLaVA-NeXT uses up to 2880 tokens).
\end{itemize}

Higher image resolutions and complex tasks further increase the computational demands on both vision encoders and language models.

\subsection{Benchmarking Baseline Models}

\begin{table}[t!]
    \centering
    \begin{adjustbox}{max width=0.95\linewidth}
    \begin{tabular}{lcccc}
        \toprule
        \textbf{Model} & \textbf{Token Usage (\#tok)} & \textbf{s/img} & \textbf{Memory (GB)} \\
        \midrule
        LLaVA-1.5-7B & 576 & 0.46 & 15 \\
        LLaVA-1.5-13B & 576 & 0.66 & 27 \\
        LLaVA-NeXT-7B & approx. 1.7--2.9K & 1.01 & 20 \\
        LLaVA-NeXT-13B & approx. 1.7--2.9K & 1.78 & 40 \\
        LLaVA-NeXT-34B & approx. 1.7--2.9K & 4.00 & 88 \\
        \bottomrule
    \end{tabular}
    \end{adjustbox}
    \caption{\textbf{Inference latency and memory costs for LLaVA models.} Tested with NVIDIA V100 GPUs.}
    \label{tab:llava_latency}
\end{table}

Table \ref{tab:llava_latency} shows resource usage during inference for LLaVA and LLaVA-NeXT models, evaluated on the DocVQA and ChartQA test sets~\citep{mathew2021docvqa,masry-etal-2022-chartqa}. The LLaVA-1.5 models demonstrate manageable computational costs, operable on a single V100 GPU. However, LLaVA-NeXT models, with up to 2.9K tokens, present significant challenges. Testing on an NVIDIA V100 32GB reveals that LLaVA-NeXT-13/34B cannot be accommodated on a single GPU. These findings emphasize the challenges of larger models, especially in resource-constrained environments.

\subsection{Existing Approaches to Efficiency}

Existing methods offer trade-offs. Sampler modules like the Perceiver resampler~\citep{alayrac2022flamingo} reduce token counts but add architectural complexity and require extra training~\citep{li2023blip2,bai2024qwenvl}. Some studies~\citep{liu2023improvedllava,dai2023instructblip} have also noted that resamplers may introduce difficulties in generating both lengthy and concise responses effectively, leading to the development of supplementary models to ensure fluent responses~\citep{bai2024qwenvl,laurencon2024matters}.
Ongoing initiatives, such as the use of convolutional or pooling layers, are also being explored~\citep{cha2023honeybee,abdin2024phi3}. We believe these concurrent developments could complement our work, demonstrating the potential for integrated use. For a more detailed discussion, please visit Section~\ref{sec:related_work} and Appendix~\ref{sec:ablation_resampler}.

Notably, improving models based on the simple architecture for enhanced speed and performance remains a high-impact research area. Due to its inherent simplicity, the LLaVA architecture is already seamlessly integrated with many popular libraries, such as \textit{SGLang}~\citep{zheng2023efficiently}, thereby facilitating broader use and easier deployment in a variety of ongoing real-world applications. With these considerations in mind, in this work, we enhance the LLaVA architecture to address existing limitations, focusing on improving performance and usability without sacrificing simplicity.

\section{Efficient Language and Vision Assistant} \label{sec:remedies}

\subsection{Preliminary: Base Architecture Modification and Initial Data Curation} \label{sec:starting_point}

To identify the most effective model architecture, we test various LLMs ranging from 160M to 13B as follows: Llama-160M, Tiny-Vicuna-1B, Phi3-3.8B, Vicuna-7B, and Vicuna-13B\footnote{The links are~\url{https://huggingface.co/Felladrin/Llama-160M-Chat-v1},~\url{https://huggingface.co/Jiayi-Pan/Tiny-Vicuna-1B},~\url{https://huggingface.co/microsoft/Phi-3-mini-4k-instruct},~\url{https://huggingface.co/lmsys/vicuna-7b-v1.5},~and~\url{https://huggingface.co/lmsys/vicuna-13b-v1.5}.}.

For the vision encoder, we replace the OpenAI CLIP-Large-336-14 module (used in LLaVA-1.5) with OpenAI CLIP-Base-224-32 and utilize the AnyRes technique~\citep{liu2024llavanext} to optimize the balance between resolution and token count. OpenAI CLIP-Large-336-14 processes a 336x336 area into 576 tokens, while OpenAI CLIP-Base-224-32 processes a 224x224 area into 49 tokens. By applying AnyRes, we increase the resolution to 896x676px, with the maximum token count capped at 637, in contrast to LLaVA-NeXT's 2880 tokens for 672x672px. It is important to note that patch size alone does not determine performance; rather, a balanced consideration with resolution ensures optimal results. Through this modification, we achieve a higher input resolution with a slight increase in maximum token count and a marginal decrease in performance (from LLaVA-1.5-7B's 40.1 to 39.8; see \S4.1.1 in Figure~\ref{fig:teaser}).

Next, we expand the dataset using Idefics2~\citep{laurencon2024matters}, LLaVAR~\citep{zhang2023llavar}, and several open-source datasets to enhance performance. This includes 1.1M images for alignment tasks and 1M for instruction tuning. Further dataset details are in Appendix~\ref{sec:hyperparameters}. As shown in Figure~\ref{fig:teaser}, this approach improves performance (39.8 to 50.7) but does not reach LLaVA-NeXT-7B's levels at 53.2 (see \S4.1.2 in Figure~\ref{fig:teaser}).

\subsection{Problem Definition and Strategies} \label{sec:problem_and_improvement}

Despite multiple optimizations, the model exhibits performance issues, particularly in generating hallucinations—incorrect responses resulting from inherent bias rather than accurate visual interpretation. See Appendix~\ref{sec:appendix_lens} for our preliminary analysis on this issue. These problems are especially critical in tasks that require strong integration of visual and textual information.

\paragraph{Hypothesized Challenges.} We hypothesize two main challenges: (1) inadequate embeddings from the vision encoder, and (2) a deficiency in the basic comprehension of text within images, which is essential for performing more complex tasks.

\paragraph{Improvement Strategies.} To address these challenges, we implement: (1) a more efficient vision encoder to enhance the quality of embeddings, and (2) a training regimen that prioritizes text comprehension before proceeding to more complex tasks.

To test our hypotheses, we conduct a series of comprehensive ablation experiments. Figure~\ref{fig:teaser} illustrates our development stages. We track the effectiveness of our model modifications using various text-centric evaluation benchmarks, including {DocVQA} (\textbf{Doc})~\citep{mathew2021docvqa}, ChartQA (\textbf{Chart})~\citep{masry-etal-2022-chartqa}, {InfographicVQA} (\textbf{Info})~\citep{Mathew_2022_WACV}, and {SEED-2-Plus} (\textbf{SD2P})~\citep{li2024seed2plus}. Additionally, we employ widely-used general multimodal benchmarks such as {SEED-IMG} (\textbf{SD-I})~\citep{li2023seed}, {MMStar} (\textbf{MMS})~\citep{chen2024we}, {ScienceQA-IMG} (\textbf{SciQA})~\citep{lu2022learn}, and {HallusionBench} (\textbf{Hall})~\citep{Guan_2024_CVPR}. Our primary objective is to enhance performance on text-centric tasks while maintaining competitive performance on general tasks and ensuring low inference costs.

In the followings, we introduce each proposed module in detail and conduct extensive ablation studies, analyzing the impact of removing each component from the final model configuration.

\input{latex/ablation_table_readclip}

\begin{figure}[t]
    \centering
    \includegraphics[width=\linewidth]{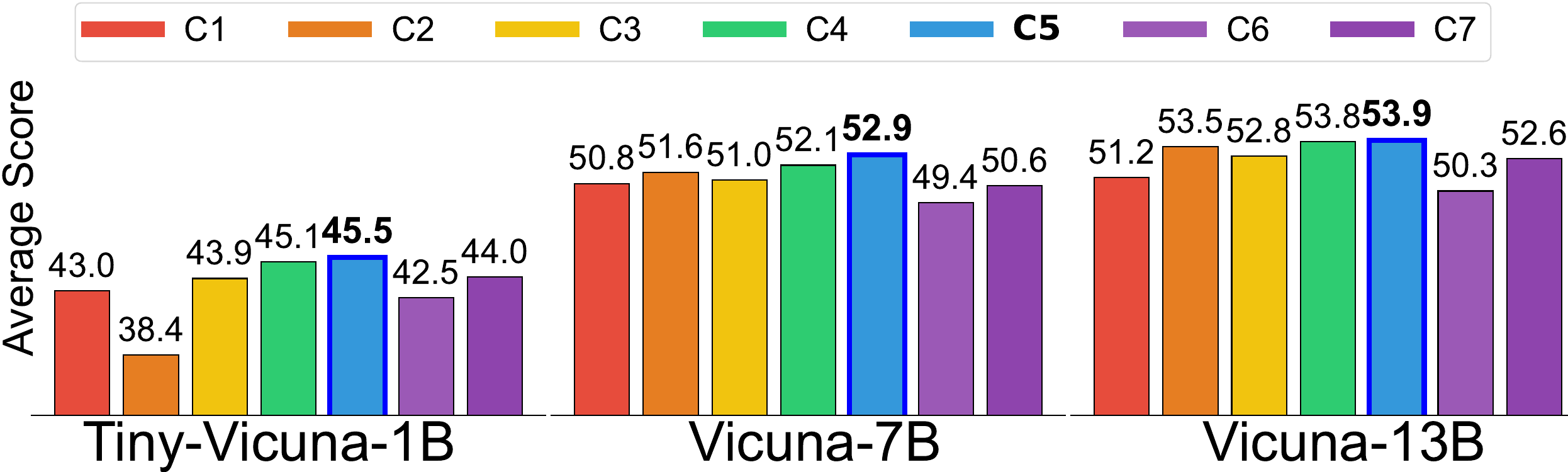}
    \caption{\textbf{Performance of various vision encoder configurations at 1B, 7B, and 13B.} Average scores for each configuration (C1 to C7) across 8 benchmarks.
    }
    \label{fig:ve_ablation_larger_scales}
\end{figure}

\subsection{Developing an Enhanced Vision Encoder with Weight Averaging} \label{sec:readclip}

To improve visually-situated text comprehension, we develop a new vision encoder optimized specifically for reading text within images.
Table~\ref{tab:ve_ablations_updated} presents ablation studies on various vision encoder configurations. Initially, we find that merely unfreezing the vision encoder during VLM training does not lead to notable performance improvements (\textbf{C2}). Next, we implement a two-step approach: (1) We first unfreeze the vision encoder and train it on a small scale VLM (1B is used) using text-centric datasets such as OCR-IDL~\citep{biten2022ocr}. This training emphasizes the \textit{Text Reading} task~\citep{kim2022donut}, where the model reads text embedded within images, allowing the vision encoder to adjust and enhance its text recognition capabilities. (2) Subsequently, we extract the enhanced vision encoder, denoted as \textit{REncoder}, from this text-centric VLM. Note that, the text-centric VLM used to derive the \textit{REncoder} is not utilized in later stages. When training a VLM with the newly obtained \textit{REncoder}, we observe significant improvements in text-centric tasks (\textbf{C3}), although its performance on general image tasks diminishes.

Now we have two specialized encoders: the original CLIP for general tasks and the \textit{REncoder}. Drawing inspiration from previous work on \textit{Weight Averaging}~\citep{pmlr-v162-wortsman22a}, we experiment with averaging the weights of the original encoder and the \textit{REncoder}. Interestingly, this approach yields promising results (\textbf{C4}). Furthermore, by slightly adjusting the weight averaging ratios to favor the \textit{REncoder}, we achieve marginally better performance on text-centric tasks (\textbf{C7}).

To further enhance robustness, we train 12 \textit{REncoder}s with different random seeds and then average their weights, a practice inspired by \citet{pmlr-v162-wortsman22a}. This averaging process, taking approximately 1.7 days on 8 V100 GPUs per phase, yields an encoder that substantially improves text comprehension while maintaining general capabilities (\textbf{C5}). More training details are in Appendix~\ref{sec:appendix_vision_encoder}.

In summary, the core idea is simple. We \textbf{(1) unfreeze the encoder and train a small VLM for text reading tasks, and retrieve the specialized encoder}, and \textbf{(2) make it robust to various tasks by applying weight averaging}. Finally, the produced vision encoder is used to build an efficient language and vision assistant, \textsc{Elva}. Our new \textbf{\textsc{Elva}-Encoder} (\textbf{C5}) brings substantial enhancements in text-centric tasks compared to the original base (\textbf{C1}). While there is still a reduction in general image performance compared to merely training LLaVA-1.5 on our data (\textbf{C6}), understanding the trade-offs in \textbf{C6} is key to fully appreciating the balance we achieve. We effectively balance overall performance and computational cost within the scope of CLIP-Base parameters (88M). The \textsc{Elva}-Encoder configurations demonstrate notable success overall, as shown in Table~\ref{tab:ve_ablations_updated} and Figure~\ref{fig:ve_ablation_larger_scales}.

\input{latex/abltion_table_rr}

\begin{figure}[t!]
    \centering
    \includegraphics[width=\linewidth]{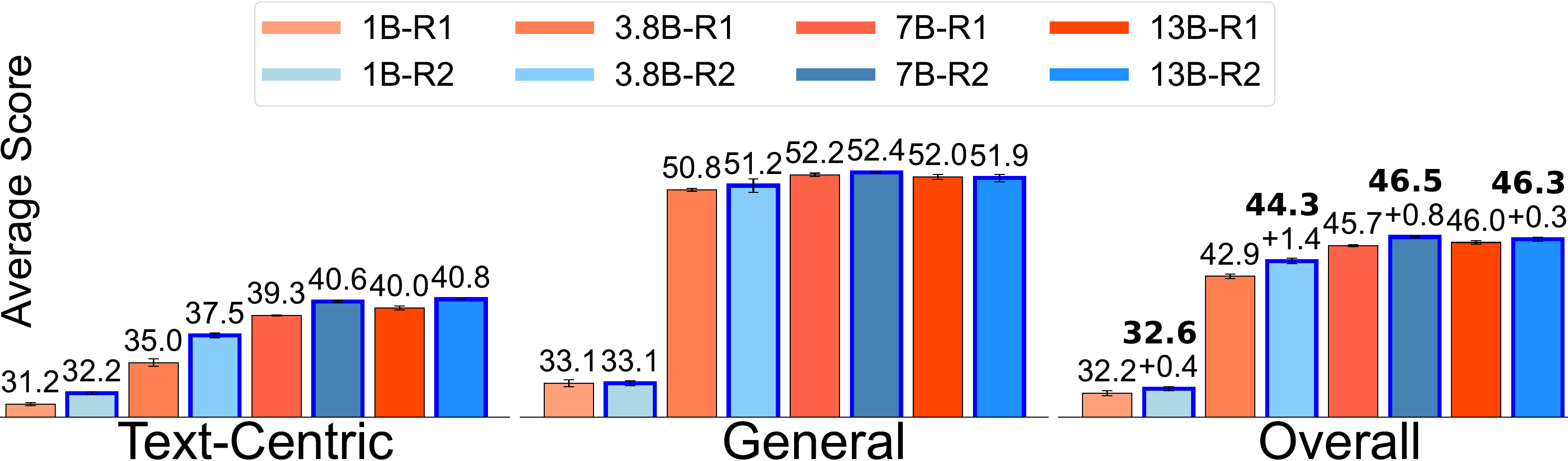}
\caption{\textbf{Impact of RR-Prompt with a 10\% dataset subset.} Results demonstrate effects during training.}
    \label{fig:rr_ablations_smalldata}
\end{figure}

\subsection{Augmenting Text Understanding in Images with Read-and-Reason Prompt}
Models like LLaVA~\citep{liu2023llava,liu2023improvedllava} utilize OCR, augmenting user queries with OCR results. However, a comprehensive investigation of methods for supervising textual information during visual instruction tuning remains underexplored.

Table~\ref{tab:rr_ablations} shows ablation studies investigating text reading tasks during visual instruction tuning. Set \textbf{R1} follows standard practices using datasets without additional text reading components. In \textbf{R2}, inspired by \textit{Prompting}~\citep{NEURIPS2020_1457c0d6}, we incorporate an initial QA task, \textit{``What is written in this image?''} before QA on text-rich images. For example, with an image of a restaurant menu, the model first \textbf{reads all text before querying} about menu items or prices. This incremental addition improves performance significantly from \textbf{R1 to R2}, especially in text-rich tasks. We annotate datasets using OCR engines for this purpose.

Our further explorations assess this approach in resource-scarce environments, using ~10\% of the original instruction tuning dataset size. Figure~\ref{fig:rr_ablations_smalldata} shows results across 1B to 13B parameter models. We also explore the supervision structure's impact by comparing ``Read-and-Reason'' versus ``Reason-and-Read'' approaches. \textbf{R3} models perform text reading last to evaluate this. Results confirm that ``Read and Reason'' is more effective, emphasizing structured prompting's importance in model learning. Lastly, we evaluate the effect of providing read text as context without explicit supervision (\textbf{R4}). Explicit supervision with text information yields marginal improvements in text-centric tasks.

In summary, the proposed core idea is to utilize Read-and-Reason Prompt (RR-Prompt) during model training to enhance text understanding in images. This approach, validated through ablation studies, shows significant performance improvements, especially in text-rich tasks. Note that the \textbf{RR-Prompt is used during training}; during inference, the model directly engages in reasoning, leveraging the enhanced capabilities acquired through the RR-Prompt, ensuring efficiency without needing an explicit text reading stage.

\subsection{Bringing It All Together}
To develop a more robust model capable of handling a wider range of tasks, we scale up the model development by incorporating diverse datasets beyond merely text-centric tasks. Our final model integrates four additional datasets: Vision-Flan~\citep{xu-etal-2024-vision}, RefCOCO~\citep{kazemzadeh-etal-2014-referitgame}, VG~\citep{DBLP:journals/ijcv/KrishnaZGJHKCKL17}, and CORD~\citep{park2019cord}. By incorporating these additional datasets, we aim to enhance both the performance and generalizability of our model. The final training involved 11K steps with a batch size of 128. The specific dataset details and schedules are in Appendix \ref{sec:appendix_implementation}. As demonstrated in Figure \ref{fig:teaser}, our final configuration shows solid performance.

\section{Experimental Assessment} \label{sec:mainexp}
\input{latex/main_table}

In this section, we rigorously test and evaluate our \textsc{Elva} models under varying conditions. We aim to understand their capabilities and limitations by benchmarking them against baseline models across both text-based and image-based tasks.

\subsection{Framework}

Our evaluation process extends beyond our initial eight datasets, utilized in our ablation studies (See \S\ref{sec:problem_and_improvement}). To further enrich our examination, we have included additional diverse datasets such as \textbf{AI2D}~\citep{Kembhavi2016ADI}, MathVista-TestMini~\citep{lu2024mathvista} (\textbf{Math}), LLaVA-Bench~\citep{liu2023llava} (\textbf{LBen}), along with Parsing-Bench (\textbf{PBen}) proposed in this work.

\subsection{Generated Scenario-Based Benchmarks}\label{sec:new_datasets}

In our research, we identify a significant gap in datasets representing real-world user scenarios for document assistants. To address this, we create the following datasets. These datasets will be open-source, and more details are in Appendix~\ref{sec:appendix_new_datasets}.

\paragraph{CORD-Instruct.} Building on the CORD dataset, which consists of Indonesian receipts and their JSON annotations, CORD-Instruct provides instructional sets for models to generate outputs in JSON, XML, or Markdown formats. We utilize the OpenAI GPT-3.5 API to create these instructional sets, ensuring the exclusion of any erratic samples.

\begin{figure}[!t]
    \centering
    \includegraphics[width=\linewidth]{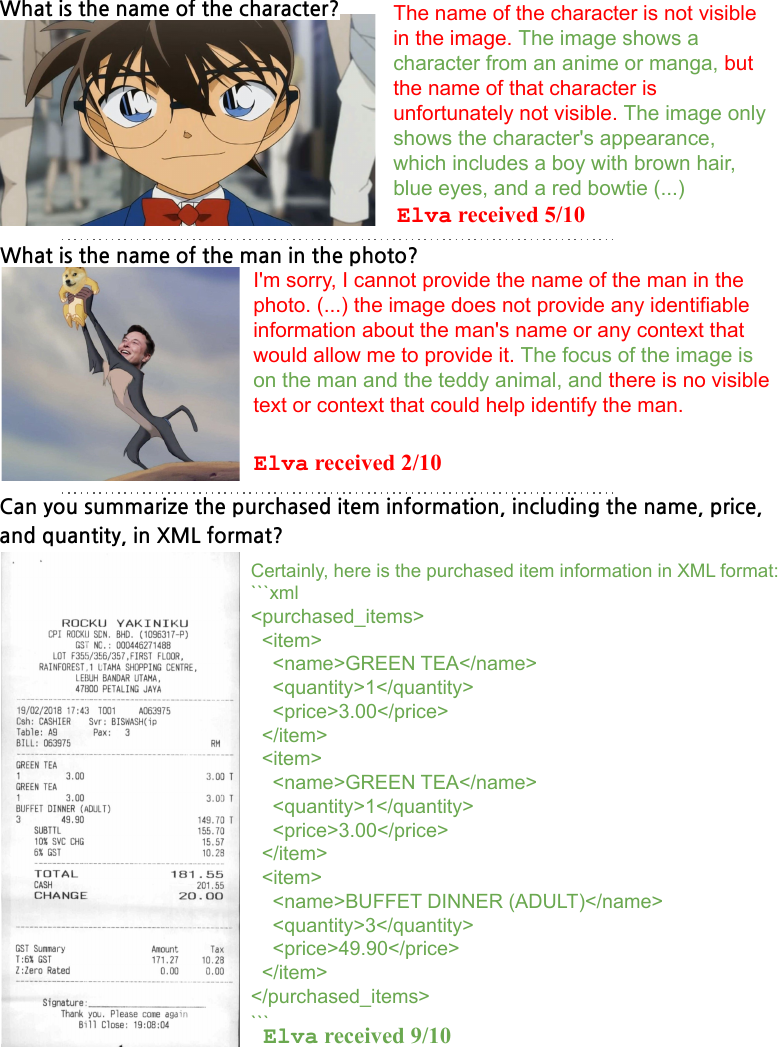}
    \caption{\textbf{Example Results of \textsc{Elva} on LLaVA-Bench and Parsing-Bench.} The strengths and weaknesses of \textsc{Elva} are illustrated.}
    \label{fig:case_study_with_llavabench_and_parsingbench}
\end{figure}

\paragraph{Parsing-Bench.} Inspired by the LLaVA-Bench and \textit{LLM-as-a-Judge}~\citep{zheng2023judging}, we develop Parsing-Bench to address the limitations of existing benchmarks like LLaVA-Bench, which include limited document-related samples and do not sufficiently reflect real user needs. Figure \ref{fig:case_study_with_llavabench_and_parsingbench} presents example cases and model predictions. To test the model's ability to extract information from new documents, we create this dataset using 30 images from Brazilian Identity Documents~\citep{sibgrapi_estendido} and SROIE~\citep{8977955}, which are not used during training.

\subsection{Results}
Table~\ref{tab:main_table} provides a comprehensive comparison of our \textsc{Elva} models against baselines such as PaliGemma~\citep{beyer2024paligemmaversatile3bvlm}, Qwen-VL~\citep{bai2024qwenvl}, and LLaVA models~\citep{liu2023llava,liu2023improvedllava,liu2024llavanext} across multiple tasks. The results, either reproduced or sourced from original papers, are validated using \textit{VLMEvalKit}~\citep{2023opencompass} and the official code by \citet{liu2023llava}.

The \textsc{Elva} models consistently exhibit strong performance on both text-centric and general multimodal benchmarks. The \textsc{Elva}-0.2B model, despite its smaller parameter count, performs commendably across various tasks. Larger models ranging from 1B to 13B demonstrate superior performance, highlighting the advantages of increased model capacity. Notably, \textsc{Elva} achieves excellent latency and memory efficiency, reinforcing its practicality for diverse real-world applications.

Although the text-centric benchmark performance is strong, there is a slight difference compared to concurrent leading specialized models~\citep{hu2024mplugdocowl}, as shown in Table \ref{tab:model_comparison}. However, our focus is on developing generalist models rather than specialized models. A detailed analysis on this is provided in Section \ref{sec:anal_sota}. Additionally, a notable limitation is observed in the LLaVA-Bench, where \textsc{Elva} models underperform compared to LLaVA models. As this dataset comprises only 24 images, interpretation requires caution. Further analysis is discussed in Section \ref{sec:case_study}.

In summary, \textsc{Elva} models demonstrate robust performance across a wide range of tasks and benchmarks. While increased model capacity generally enhances performance, efficiency and latency considerations are essential for practical deployment. Our main results highlight \textsc{Elva}'s efficiency and balanced performance, underscoring the contributions and objectives of our study.

\section{Further Analyses and Discussions} \label{sec:anal}

\begin{table}[t!]
  \centering
  \begin{adjustbox}{width=\linewidth}
  \setlength\tabcolsep{5pt} 
  \renewcommand{\arraystretch}{1.1} 
    \begin{tabular}{l|ccc|ccccc}
      \toprule
      \textbf{Model} & \textbf{Doc} & \textbf{Chart} & \textbf{SD2P} & \textbf{SD-I} & \textbf{MMS} & \textbf{SciQA} & \textbf{Hall} & \textbf{LBen} \\ 
      \midrule
      DocOwl1.5-8B & \underline{81.6} & \textbf{70.7} & 50.2 & 50.2 & 34.7 & 65.7 & 28.9 & 35.3 \\
      DocOwl1.5-8B-Chat & \textbf{82.2} & \underline{69.6} & \underline{52.4} & 50.9 & 34.4 & 65.0 & 30.4 & 39.5 \\
      \rowcolor{maroon!15} \textbf{Elva-7B (ours)} & 69.1 & 61.8 & 47.7 & \underline{62.6} & \underline{35.4} & \underline{74.7} & \textbf{56.8} & \underline{50.7} \\
      \rowcolor{maroon!15} \textbf{Elva-13B (ours)} & 71.7 & 65.2 & \textbf{52.6} & \textbf{65.3} & \textbf{37.9} & \textbf{77.7} & \textbf{56.8} & \textbf{51.0} \\ 
      \bottomrule
    \end{tabular}
  \end{adjustbox}
\caption{\textbf{Comparison with Specialized VLMs.} \textsc{Elva} shows the balanced scores on diverse benchmarks.}
  \label{tab:model_comparison}
\end{table}

\subsection{Comparisons with Specialized Models} \label{sec:anal_sota}
Given the rapid evolution in this field, evaluating our model against recent advancements is vital to highlight our contributions. We compare our work with mPLUG-DocOwl1.5~\citep{hu2024mplugdocowl}, one of the concurrent state-of-the-art models. Since their results on general multimodal benchmarks are not available, we measure the performances, ensuring accuracy with a sanity check on ChartQA. The results in Table~\ref{tab:model_comparison} suggest that while mPLUG-DocOwl1.5 excels in text-rich document VQA, it faces challenges in general multimodal tasks. This underscores our focus on developing a generalist model that balances task proficiency and broader efficiency.

\subsection{Ablations with LLaVA-NeXT Variants}

\input{latex/ablation_table_llavanext}

To test the impact of reducing the number of tokens in LLaVA-NeXT models, we constraine the grid size, resulting in a maximum token count of 1728 (either 336x672 or 672x336 pixels). 
As shown in Table~\ref{tab:ablation_table_llavanext}, reducing the vision token count leads to significant performance drops across all evaluated tasks. For example, the performance of the 13B model on DocVQA decreases from 69.8 to 53.9 when the token count is restricted, with similar trends observed in other variants.
This analysis highlights the trade-off between token count and model performance: while reducing tokens can enhance computational efficiency, it may lead to a compromise in accuracy.

In contrast, \textsc{Elva} models demonstrate strong performance along with improved efficiency in both speed and memory usage, underlining their robustness in handling high-resolution text-centric tasks efficiently.
The \textsc{Elva} models effectively balance performance and efficiency, outperforming the LLaVA-NeXT variants with reduced token counts.

\subsection{Discussion on Memory and Time Costs}

\begin{figure}[t!]
    \centering
    \includegraphics[width=\linewidth]{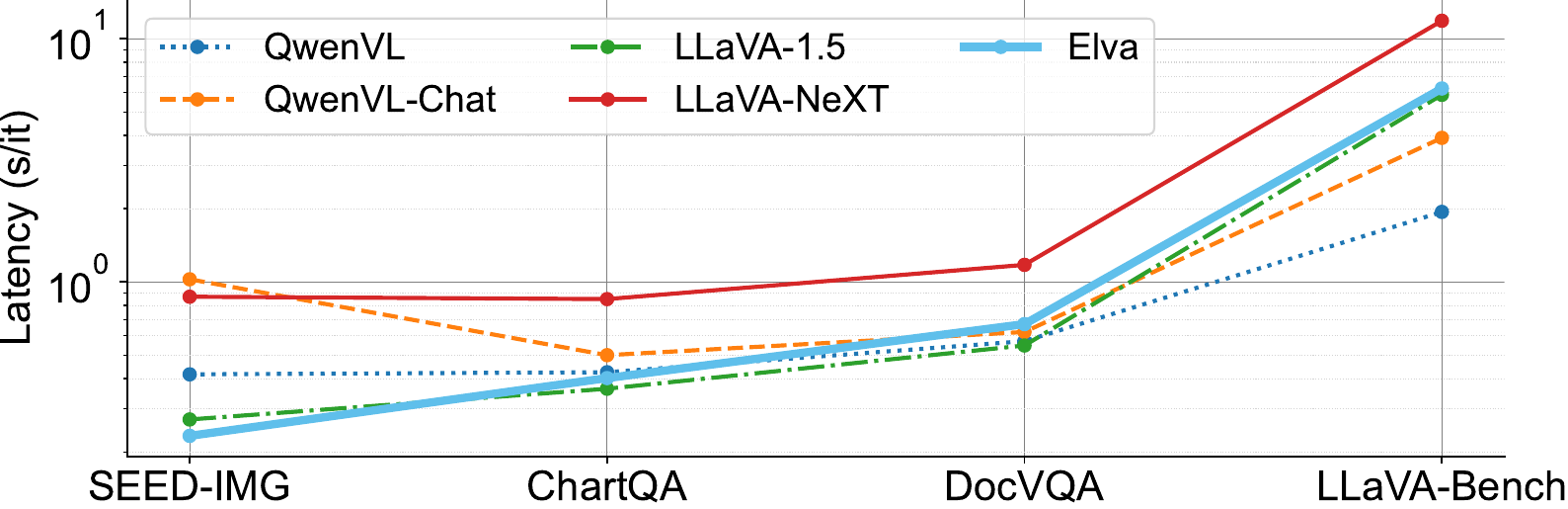}
    \caption{\textbf{Latency comparison across multiple benchmarks.} \textsc{Elva} delivers promising results.}
    \label{fig:ablations_latency}
\end{figure}

We evaluate latency for ChartQA and DocVQA, as these tasks relate closely to real-world document information extraction scenarios and offer user-centric metrics. Multiple-choice evaluations like SD-I are less indicative of actual user scenarios. Benchmarks requiring longer answers, such as LLaVA-Bench, show inconsistent results due to varied answer lengths across models. Consequently, we focus on ChartQA and DocVQA for latency assessments but also include SD-I and LLaVA-Bench results in Figure~\ref{fig:ablations_latency}. These findings indicate \textsc{Elva} maintains promising latency across varied contexts.

While this study primarily focuses on inference time costs, training costs are equally important for practitioners. Despite handling large datasets, our lightweight vision encoding ensures high training throughput. With training time costs 1.32 to 1.78 times that of LLaVA-NeXT, we find this a fair trade-off for the gains in efficiency and performance. Our approach remains competitive and more resource-efficient compared to several contemporary models. For detailed analysis, please refer to Appendix~\ref{sec:model_training_time_cost_appendix}.

Regarding memory usage, practical deployment often uses quantization~\citep{dettmers2023qlora}, significantly reducing memory costs. For instance, the LLaVA-NeXT-13B model originally requires two V100 GPUs but can run on a single V100 with quantization, albeit with more latency and some performance loss. Despite these drawbacks, quantization shows promise and will likely improve. Our \textsc{Elva} models, built for efficiency, complement these advancements, promising even greater value when combined with quantization techniques.

\subsection{Case Study on LLM-as-a-Judge Sets}\label{sec:case_study}

\textsc{Elva} models perform robustly across benchmarks, excelling in text-centric tasks, but show performance dips on LLaVA-Bench. These issues often arise from entity recognition failures, leading to penalized scores despite the response being logically sound. For example, when asked to name an anime character, \textsc{Elva} scores 0.5 for failing to provide the correct name despite a detailed description. Similarly, it scores 0.2 in identifying a person in a photo without textual clues. These dips highlight limitations in the \textsc{Elva}-Encoder's (88M) entity memorization capacity and its reliance on sufficient information present in images. Adding targeted data to memorize entities during instruct tuning might address these issues, though feasibility is debatable. On the other hand, \textsc{Elva} excels in tasks like summarizing receipt data into XML format, scoring 0.9, demonstrating its ability to accurately read, comprehend, and organize text information. 
For more discussion and analysis on this, refer to Section \ref{sec:limitation} and Appendix \ref{sec:appendix_further_lben}.

\subsection{Discussion on Leveraging OCR}

Incorporating OCR can be an effective option for handling text-rich high-resolution images~\citep{kim-etal-2023-visually}.
When OCR outputs are incorporated as contextual information during inference, as demonstrated in Table \ref{tab:ablation_table_withocr}, notable enhancements are observed, particularly benefiting \textsc{Elva}.
However, OCR processing has costs. Using the CLOVA OCR API\footnote{\url{https://clova.ai/ocr/en}}, our tests on the DocVQA dataset average about 4 seconds per sample. Faster OCR engines exist but often at the expense of accuracy. Additionally, upscaling VLMs to handle very high resolutions (e.g., 4K, 8K) may not be practical. Thus, leveraging OCR and similar tools remains a valuable area of exploration, aiming to balance specialized tools and VLMs for optimal performance.

\input{latex/ablation_table_withocr}

\section{Related Work}\label{sec:related_work}

\paragraph{Visually-Situated Natural Language Understanding (NLU).}
Visually-situated NLU requires detailed image comprehension and high-resolution processing. Initial VLMs relied on OCR for text extraction. For instance, \citet{xu2020layoutlm} utilizes OCR and integrates textual and layout information for document understanding. The field then moved to OCR-free methods~\cite{kim2022donut,10.1007/978-3-031-41498-5_16,lee2023pix2struct}, with models like Donut~\citep{kim2022donut} enabling efficient visually-situated NLU. 

\paragraph{Multimodal LLMs (MLLMs).}
MLLMs enhance multimodal comprehension by utilizing LLMs' language understanding. Early models, such as LLaVA~\cite{liu2023llava} and BLIP-2~\citep{li2022blip}, align visual representations with frozen LLMs. More recent models, such as LLaVA-1.5~\citep{liu2023improvedllava}, LLaVA-NeXT~\citep{liu2024llavanext}, and Qwen-VL~\citep{bai2024qwenvl}, unfreeze LLM parameters and use extensive resources~\cite{laurencon2024matters,laurençon2024building,internlmxcomposer2_4khd,hu2024mplugdocowl}. Studies like MM1~\citep{mm1-methods-analysis-insights} provide thorough architecture ablations but focus less on visually-situated NLU, and without offering code or model weights. \citet{laurencon2024matters} conduct multiple ablations and make their methods more accessible to the public, but focus less on reducing resource costs.
Small VLMs with under 3B scales~\citep{chu2024mobilevlm} are emerging. However, there is still a need for compact VLMs for tasks like high-resolution document image processing, emphasizing our work's focus and contribution.

\paragraph{MLLMs for Visually-Situated NLU.}
Early efforts integrated OCR for text-heavy inputs~\citep{liu2023llava,li2022blip,liu2023improvedllava}, but there is a shift to OCR-free designs~\citep{kim-etal-2023-visually,liu2024llavanext,laurencon2024matters,laurençon2024building,internlmxcomposer2_4khd,hu2024mplugdocowl}. High-performance models use increased input resolutions, raising costs. For instance, LLaVA-NeXT~\citep{liu2024llavanext} uses 2880 tokens for 672x672 pixels, leading to high costs, furthered by later models~\citep{internlmxcomposer2_4khd,wang2024qwen2vlenhancingvisionlanguagemodels}. \citet{internlmxcomposer2_4khd} scaled input resolution to 3840×1600 pixels, requiring over 8K tokens.

\paragraph{MLLMs with Vision Token Sampling.}
To reduce token usage, studies have explored vision token sampling techniques. For example, Qwen-VL~\citep{bai2024qwenvl} and Idefics2~\citep{laurencon2024matters} utilize the Perceiver resampler~\citep{alayrac2022flamingo}. These models, however, entail high training costs: Qwen-VL uses over 1.4B data points and Idefics2 over 1B, contrasting with smaller data usage in the LLaVA series~\citep{liu2023llava,liu2023improvedllava,liu2024llavanext}. Some studies~\citep{liu2023improvedllava,dai2023instructblip} have found that models using resamplers can face challenges in generating both lengthy and brief responses, leading to the development of additional models like Qwen-VL-Chat~\citep{bai2024qwenvl} and Idefics2-Chatty~\citep{laurencon2024matters}. Simpler approaches, such as using convolutional or pooling layers, have also been explored~\citep{cha2023honeybee,abdin2024phi3}. These methods align orthogonally with the proposed methods in this paper, underscoring the potential for combined use.

Overall, further research is needed to identify key factors for effective VLM design. For additional comparisons, please refer to Appendix~\ref{sec:model_training_time_cost_appendix}.

\section{Conclusion}
This study introduces \textsc{Elva}, a robust and efficient model framework for diverse multimodal tasks, including visually-situated NLU. Empirical results demonstrate that \textsc{Elva} surpasses existing baselines, with notable memory and latency efficiency. Comprehensive experiments and analyses identify key components driving \textsc{Elva}'s enhanced performance. Additionally, our analysis highlights both strengths and limitations, offering insights for further development. We envision our approach extending to other domains and tasks, particularly those requiring high-resolution and visually-situated NLU, even in resource-constrained environments.

\section{Limitations} \label{sec:limitation}

Despite the significant advancements demonstrated by \textsc{Elva}, several limitations remain. Firstly, \textsc{Elva} occasionally struggles to recognize specific entities within images, leading to reduced accuracy in responses, even when they are logically sound. This suggests that the vision encoder may have limitations in recognizing long-tail entities, highlighting the need for further analyses and future research.

Managing very high-resolution images (4K or 8K) is still challenging. While the proposed methods advance the handling of such images, they are not sufficient for easy processing beyond this resolution. We should continue to balance performance improvements with computational resource requirements. For high-resolution document images, incorporating OCR could be a viable option, but it introduces latency and potential accuracy trade-offs, necessitating additional research.

Although \textsc{Elva} achieves lower inference costs and maintains reasonable training times, processing large data volumes can lead to moderate time differences. As discussed in Appendix~\ref{sec:model_training_time_cost_appendix}, we have made significant improvements with acceptable increases in training costs, but ongoing optimization in both training efficiency and performance remains necessary.

Future research should focus on enhancing entity recognition, improving training efficiency, and refining OCR integration. Exploring the balance between specialized tools like OCR and an end-to-end VLM is crucial for optimizing performance. Additionally, expanding \textsc{Elva}'s capabilities to handle multilingual or video tasks would further increase its applicability and utility.

\section{Ethical Considerations}

Developing \textsc{Elva} involves important ethical responsibilities such as reducing data biases and ensuring transparency. To manage these, we use only controlled and verified open-source datasets for model training. Currently, we rely on the autoregressive models' direct output, but we could also use post-processing techniques or additional training methods to address biases and privacy issues better. By open-sourcing our models and datasets, we encourage peer reviews and collaboration to solve ethical challenges, promoting accountability. These steps help ensure that \textsc{Elva} upholds high ethical standards and is used for beneficial purposes while minimizing risks.

\section*{Acknowledgements}

We extend our sincere gratitude to Bado Lee, Daehee Kim, Taeho Kil, and Hodong Lee for their meticulous proofreading of this manuscript. Their input greatly increased its clarity and coherence. We are also immensely grateful to our colleagues in the NAVER Cloud Hyperscale AI Vision Understanding Team and KAIST AI LKLab. Their constant support and encouragement have been a great source of motivation throughout this work. This work was partly supported by KAIST-NAVER Hypercreative AI Center.

\bibliography{anthology,custom}

\appendix

\section{Additional Analysis and Comparison}\label{sec:model_training_time_cost_appendix}

\subsection{Model Training Cost and Data Efficiency}
Table~\ref{tab:training_cost_appendix} provides a detailed analysis of the training costs associated with different model sizes of \textsc{Elva}, using 8 NVIDIA A100 80GB GPUs. According to estimates from the official LLaVA-NeXT blog\footnote{\url{https://llava-vl.github.io/blog/2024-01-30-llava-next}}, our models take approximately 1.32 to 1.78 times longer to train. The blog reports that it takes 20 hours to train a 7B model with 8 A100 GPUs and 24 hours for a 13B model using 16 A100 GPUs. While training times can vary depending on the testing environment, our data shows that \textsc{Elva}'s training duration results in a moderate increase that remains within reasonable expectations.

This observation becomes clearer when \textsc{Elva} is compared to other contemporary, data-intensive models. For instance, Qwen-VL~\citep{bai2024qwenvl} requires 1.4B data points for pretraining and 50M data points for instruction tuning, whereas \textsc{Elva} demonstrates a more moderate yet effective use of resources. Similarly, models such as Shikra~\citep{chen2023shikra}, Idefics2~\citep{laurencon2024matters}, and InternLM-XComposer2-4KHD~\citep{internlmxcomposer2_4khd} illustrate varying scales of resource utilization, with Shikra using 600K data points for alignment and 5.5M for instruction tuning, Idefics2 achieving results with over 1B data points, and InternLM-XComposer2-4KHD demonstrating scalability with a massive dataset and more than 8K input tokens.

Our observations in Section~\ref{sec:readclip} further reinforce the argument for \textsc{Elva}'s efficiency. Despite the extended data, LLaVA-1.5 failed to surpass the overall score we achieved with \textsc{Elva} (\textbf{C5} vs. \textbf{C6}), corroborating the efficiency and effectiveness of \textsc{Elva}. Furthermore, it is crucial to highlight the importance of inference cost. Models aiming for reduced inference costs often face expensive training costs and challenges in maintaining instruction-following capabilities across varied response lengths~\citep{dai2023instructblip,liu2023improvedllava,laurencon2024matters}. Thus, \textsc{Elva} emerges as a quick, lightweight, and cost-effective alternative within LLaVA-like simple architectures.

\begin{table}[t!]
\centering
\begin{adjustbox}{max width=\linewidth}
\begin{tabular}{l|ccc}
    \toprule
    \textbf{Base Model} & \textbf{Alignment Time} & \textbf{Instruct Tuning Time} & \textbf{Total Time} \\
    \midrule
    Llama-160M & 0.5 hours & 3.5 hours & 4 hours \\
    Tiny-Vicuna-1B & 1.5 hours & 6 hours & 7.5 hours \\
    Phi3-3.8B & 4.5 hours & 16.5 hours & 21 hours \\
    Vicuna-7B & 6.5 hours & 29 hours & 35.5 hours \\
    Vicuna-13B & 11 hours & 52.5 hours & 63.5 hours \\
    \bottomrule
\end{tabular}
\end{adjustbox}
\caption{\textbf{Training times for various model sizes on 8 A100 GPUs.}}\label{tab:training_cost_appendix}
\end{table}

\subsection{Ablations with AnyRes and Resampler} \label{sec:ablation_resampler}

As discussed in Section \ref{sec:related_work}, previous research~\citep{liu2023improvedllava,dai2023instructblip} highlights several limitations associated with resampler-based techniques. To fully understand the limitations, it is essential to empirically investigate them. This section presents our additional experiments on the effectiveness of the Perceiver Resampler~\citep{alayrac2022flamingo}, a tool commonly used in many MLLMs to reduce vision token counts~\citep{cha2023honeybee}. We conduct these experiments using the Vicuna-7B model.

For this experiment, we begin by training models using CLIP-Large-336-14, as employed in LLaVA-1.5~\citep{liu2023improvedllava}. We then introduce AnyRes~\citep{liu2024llavanext}, which can be interpreted as training the LLaVA-NeXT architecture on our same dataset and training parameters. Finally, we apply the Perceiver Resampler in an attempt to reduce the token count.

Our findings are summarized in Table~\ref{tab:perceiver_ablation_results}. The results suggest that performance is notably constrained. This limitation likely arises from the disparity in resources and data used during our compact \textsc{Elva} training, which may not be sufficient for the resampler to fully realize its potential. Increasing the dataset size and training steps might enhance the effectiveness of the resampler. Additionally, as discussed in Section \ref{sec:related_work}, several improved resampling methods are emerging, and combining them with our approach would likely yield better results.

\begin{table}[t]
    \centering
    \begin{adjustbox}{max width=\linewidth}
    \begin{tabular}{l|ccc|ccccc}
        \toprule
        \textbf{Configuration} & \textbf{Chart} & \textbf{SD2P} & \textbf{SD-I} & \textbf{MMS} & \textbf{SciQA} & \textbf{Hall} & \textbf{AI2D} & \textbf{Math} \\
        \midrule
        \rowcolor{maroon!15} \textbf{\textsc{Elva}-88M} & 61.8 & 47.7 & 62.6 & 35.4 & 74.7 & 56.8 & 66.2 & 36.6 \\
        \midrule
        \midrule
        CLIP-L-300M & 43.9 & 47.3 & 66.8 & 38.2 & 80.3 & 55.6 & 68.9 & 35.6 \\
        \:\: + AnyRes & 61.6 & 53.8 & 67.9 & 39.8 & 77.3 & 55.8 & 66.8 & 37.8 \\
        \rowcolor{redd!15} \:\: \:\: \textbf{+ Resampler} & 18.7 & 38.2 & 43.1 & 30.2 & 72.4 & 50.9 & 62.8 & 28.5 \\
        \bottomrule
    \end{tabular}
    \end{adjustbox}
\caption{\textbf{Performance evaluation with AnyRes and Perceiver resampler.} This table illustrates the performance of our proposed \textsc{Elva}-Encoder (88M) and OpenAI CLIP-Large-336-14 (300M) configurations with AnyRes and Resampler optimizations across diverse tasks. All configurations are trained using the same alignment and visual instruction tuning schedule to guarantee consistent evaluation conditions. For the \textsc{Elva} setting, the vision token ranges are 98-637, 576 for CLIP-Large; for AnyRes, it is expanded to 1728-2880, and with Resampler, it ranges between 432-720 tokens.}
\label{tab:perceiver_ablation_results}
\end{table}

\subsection{Additional Model Variants}

In this paper, we employ different LLM families across each scale. We adopt Vicuna as our base model to enable fair comparisons with the LLaVA family~\citep{liu2023llava,liu2023improvedllava,liu2024llavanext}. While more advanced LLMs like LLaMA-3\footnote{\url{https://ai.meta.com/blog/meta-llama-3}} could potentially achieve better scores, reaching state-of-the-art benchmarks is not our primary goal. Therefore, we stick with the Vicuna family for consistency. For other scales, we select LLMs that have garnered attention in recent open-source VLM projects. Tiny-Vicuna and LLaMA-Chat are chosen because they are fully open-source models, developed transparently by academic practitioners with limited resources.

We acknowledge that using LLMs trained on the same corpus could provide additional insights into scaling effects. If all these models are trained using different datasets or regimes, it may complicate the evaluation setup, making it difficult to precisely identify performance-related issues. To address this, we also train more variants using the OpenELM family~\citep{mehta2024openelmefficientlanguagemodel}, which has been recently released as open-source models. OpenELM offers transparent training details, which help elucidate scaling effects more clearly. Table \ref{tab:openelm} presents the results, demonstrating that our proposed \textsc{Elva} model scales effectively and consistently across various LLM architectures, thereby validating the robustness of our approach.

\begin{table}[t]
    \centering
    \begin{adjustbox}{max width=\linewidth}
    \begin{tabular}{l|ccc|ccccc}
        \toprule
        \textbf{Base LLM} & \textbf{Chart} & \textbf{SD2P} & \textbf{SD-I} & \textbf{MMS} & \textbf{SciQA} & \textbf{Hall} & \textbf{AI2D} & \textbf{Math} \\
        \midrule
        LLaMA-160M & 50.3 & 31.4 & 37.8 & 31.5 & 39.0 & 48.1 & 31.0 & 27.0 \\
        OpenELM-270M & 54.4 & 32.4 & 45.0 & 30.9 & 46.2 & 46.9 & 34.8 & 29.5 \\
        OpenELM-450M & 56.8 & 35.4 & 50.4 & 31.8 & 62.3 & 51.5 & 44.0 & 29.1 \\
        Tiny-Vicuna-1.1B & 57.7 & 36.9 & 52.3 & 32.6 & 63.3 & 50.4 & 46.9 & 31.7 \\
        OpenELM-1.1B & 59.3 & 40.1 & 57.1 & 31.7 & 67.8 & 50.3 & 54.4 & 33.7 \\
        \bottomrule
    \end{tabular}
    \end{adjustbox}
\caption{\textbf{Performance results across different LLM variants.} The results demonstrate the scalability and consistency of our proposed \textsc{Elva} model across different architectures.}
\label{tab:openelm}
\end{table}

\begin{table}[t!]
    \centering
    \begin{adjustbox}{max width=\linewidth}
    \begin{tabular}{lccccc}
        \toprule
        \textbf{Configuration} & \textbf{Size} & \textbf{Text-Centric} & \textbf{General} & \textbf{Overall} & \textbf{LLaVA-Bench} \\
        \midrule
        \textbf{C1.} CLIP-B-Anyres & 88M & 44.0 & 58.5 & 51.2 & 51.1 \\
        \rowcolor{maroon!15} \textbf{C5.} \textbf{Elva-Encoder} & 88M & 50.4 & 57.4 & \textbf{53.9} & 47.3 \\
        \textbf{C6.} CLIP-Large & 300M & 39.6 & 60.9 & 50.3 & 69.1 \\
        \bottomrule
    \end{tabular}
    \end{adjustbox}
    \caption{\textbf{Performance comparison across different vision encoder configurations.} The table shows the text-centric, general, and overall benchmark scores, as well as the LLaVA-Bench scores.}
    \label{tab:entity_memory}
\end{table}

\subsection{Further Analysis on LLaVA-Bench} \label{sec:appendix_further_lben}

In Section \ref{sec:case_study}, we examine the lower performance of the \textsc{Elva} model on the LLaVA-Bench. Given the small number of model parameters in the \textsc{Elva}-Encoder, we hypothesize that its ability to memorize entities might be limited, potentially contributing to its lower performance. While \textsc{Elva} often provides logically sound responses to user queries, it sometimes fails to recall specific entity names—a situation comparable to humans struggling to remember the name of an unfamiliar animated character without any contextual clues.

To investigate this further, we consider whether increasing model scale can enhance memorization capacity. We revisit our ablation models from Section \ref{sec:readclip}, focusing on the 13B models shown in Figure \ref{fig:ve_ablation_larger_scales}. Table \ref{tab:entity_memory} presents additional evaluations on the LLaVA-Bench using these models.

The results indicate that CLIP-Large performs exceptionally well on the LLaVA-Bench. However, both smaller encoder settings, \textbf{C1} and \textbf{C5}, face challenges on this benchmark. Despite these challenges, we find the Elva-Encoder to be effective in many other scenarios.

Through this analysis, we recognize the benefits of larger encoders. However, whether increasing parameters significantly enhances memorization is still an open question for future research. We also question whether VLMs should even prioritize memorizing all entities. Our work lays a solid foundation for exploring the trade-offs between model scalability and performance.

\subsection{Preliminary Analysis on Hallucinations} \label{sec:appendix_lens}

\begin{figure}[!t]
    \centering
    \includegraphics[width=\linewidth]{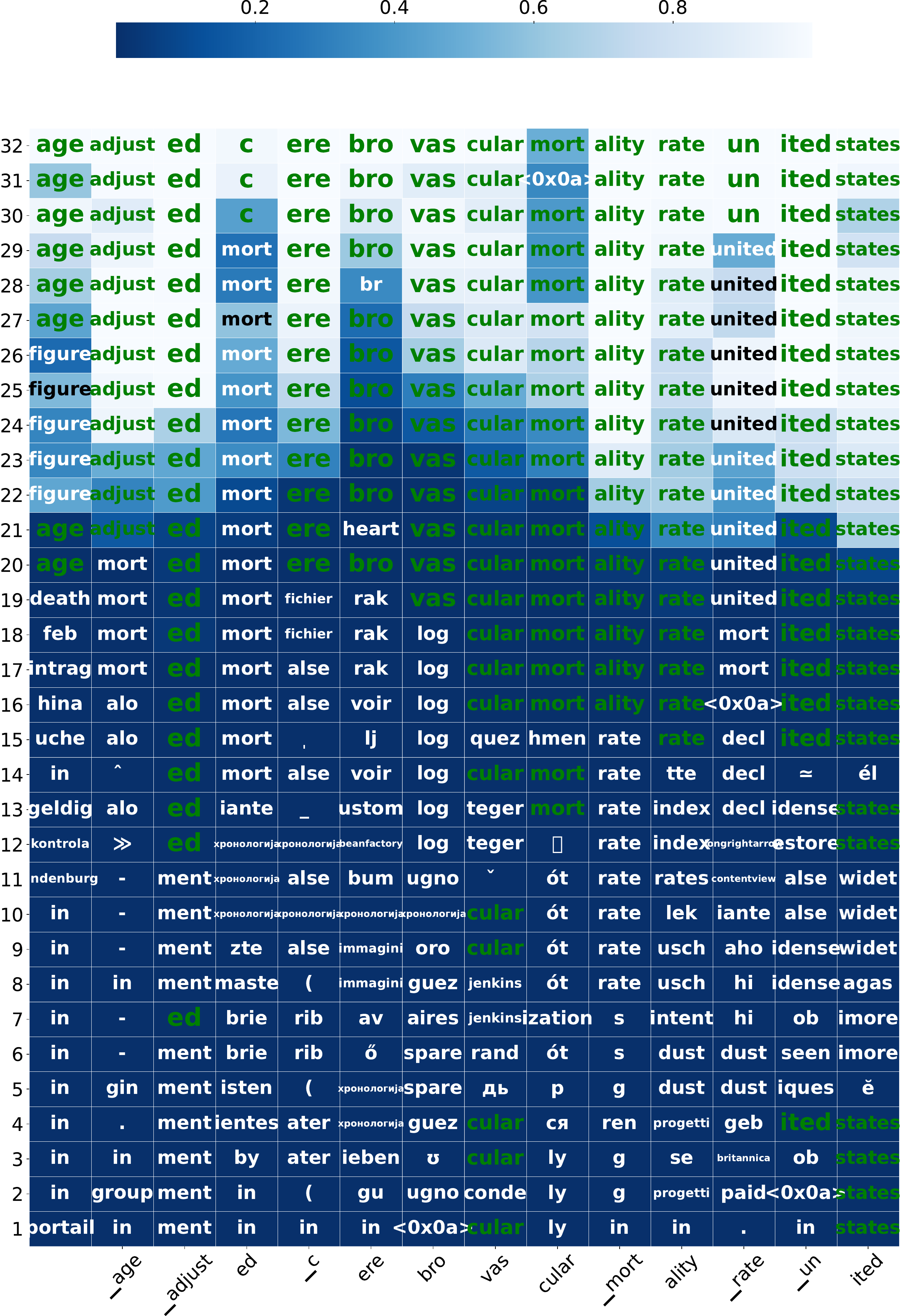}
    \caption{\textbf{Results from the \textsc{Elva} model.} The model accurately predicts the correct answer, with the correct token emerging early in the processing layers, highlighting effective vision and text integration.}
    \label{fig:lens_elva}
\end{figure}

\begin{figure}[!t]
    \centering
    \includegraphics[width=0.974\linewidth]{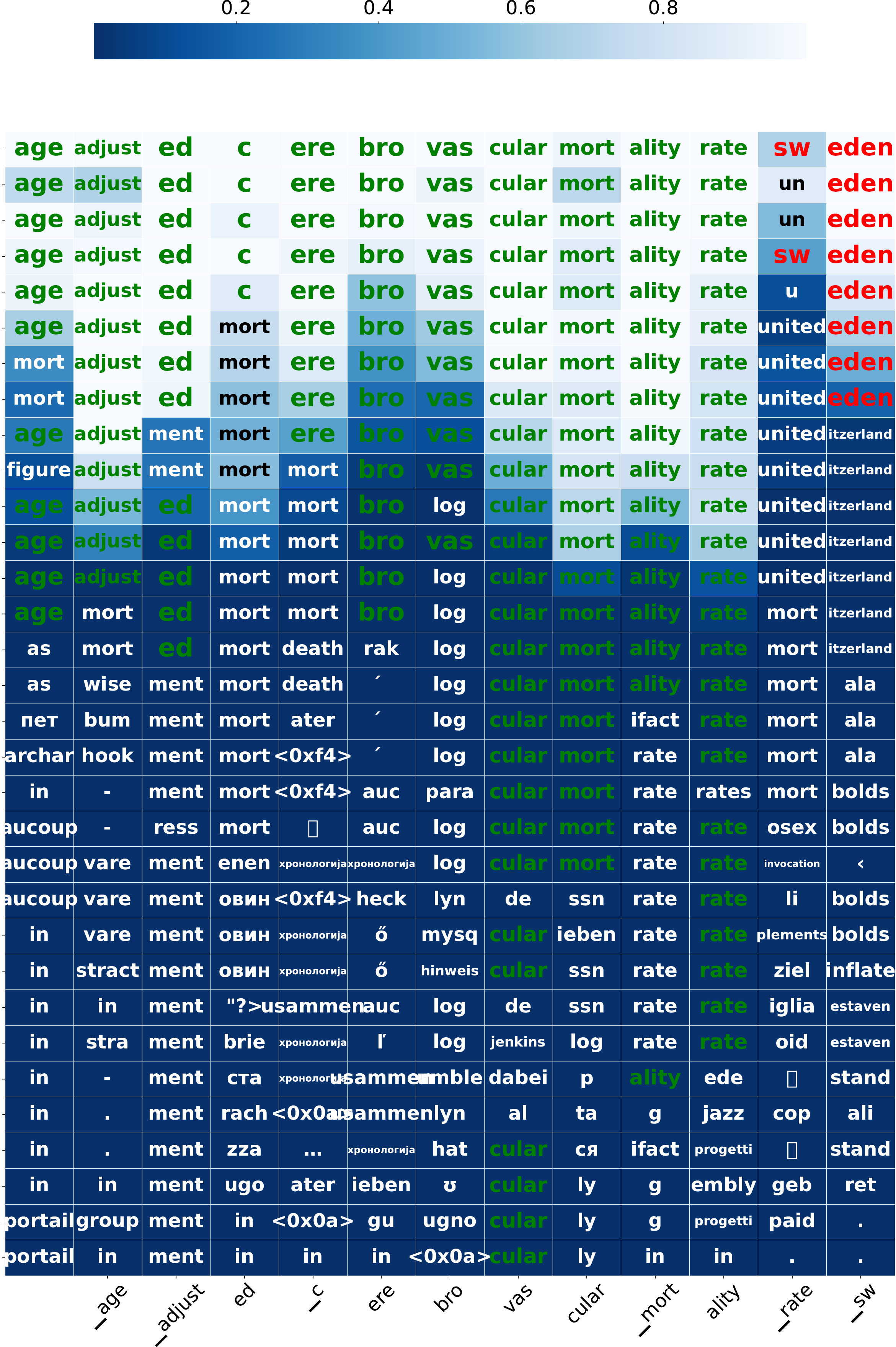}
    \caption{\textbf{Results from the ablated model.} This model incorrectly predicts \textit{``sweden''} as the answer, demonstrating the challenges faced without the \textsc{Elva}-Encoder and RR-Prompt enhancements.}
    \label{fig:lens_no_elva}
\end{figure}

\begin{figure}[!t]
    \centering
    \includegraphics[width=\linewidth]{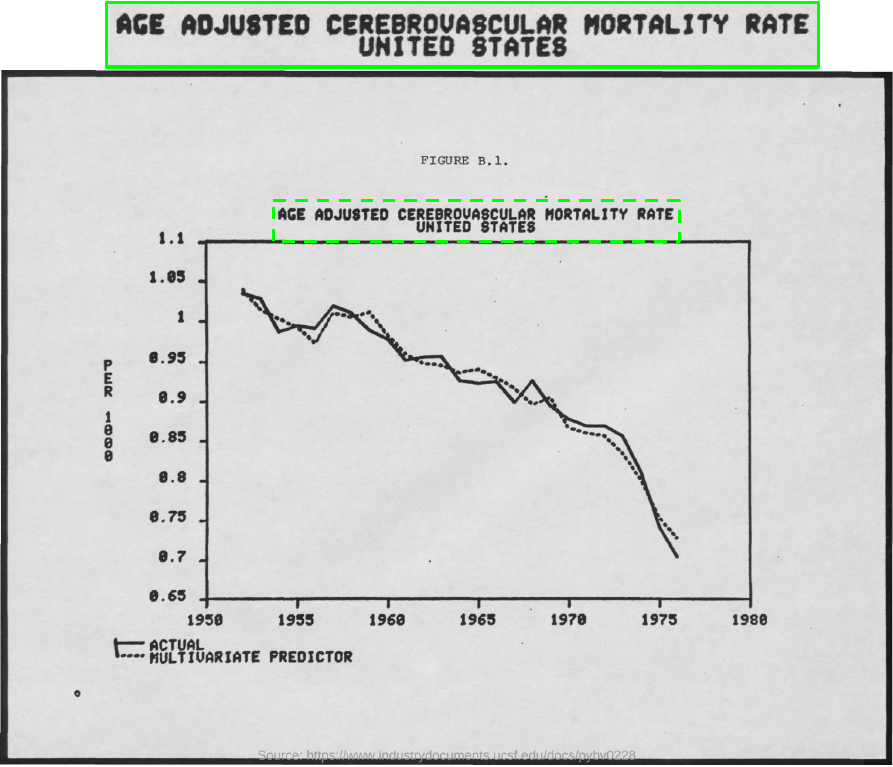}
    \caption{\textbf{DocVQA sample.} The question posed is, \textit{``What is the title of the plot?''} The model received instructions to respond concisely in lowercase. For the query, \textit{``age adjusted cerebrovascular mortality rate united states''} is the expected answer.}
    \label{fig:lens_problem}
\end{figure}

As demonstrated in Section \ref{sec:starting_point}, we explore increasing input resolution without significantly raising inference costs by integrating AnyRes into the LLaVA-1.5 model, thus avoiding excessive growth in vision token counts. Additionally, we enhance performance by expanding the training dataset. Despite these improvements, as discussed in Section \ref{sec:problem_and_improvement}, the model still faces performance issues, particularly in generating hallucinations—incorrect responses due to inherent bias rather than accurate visual interpretation. This section delves into our preliminary analysis, presenting and explaining samples that illustrate these issues.

Without applying the proposed modules, the ablation model sometimes produces unexpectedly incorrect responses in tasks that require interpreting text within images. However, it's important to note that this model, which excludes the Elva-Encoder and RR-Prompt, is not inherently inadequate. In fact, as shown in Figure \ref{fig:teaser}, its overall performance significantly surpasses that of LLaVA-1.5 (improving from 40.1 to 50.7). This underscores the significant impact of our methods. Figures~\ref{fig:lens_elva}, \ref{fig:lens_no_elva}, and \ref{fig:lens_problem} illustrate results from our \textsc{Elva} model at the 7B scale, an ablation model without the \textsc{Elva}-Encoder and RR-Prompt, and a DocVQA sample used for analysis. To better understand the model’s internal processes, we employ the \textit{Logit Lens} technique from \citet{logitlens} to visualize the behavior across all model layers in this study.

As seen in Figure~\ref{fig:lens_elva}, the Elva model accurately identifies the correct answer, with the correct token emerging as the top candidate relatively early in the processing layers. For simplicity, this analysis did not differentiate between uppercase and lowercase letters. On the other hand, Figure~\ref{fig:lens_no_elva} presents an intriguing result where the model becomes confused among various country names and incorrectly outputs \textit{``sweden''} as the answer. Notably, Figure~\ref{fig:lens_problem} shows that there is no indication in the image that resembles ``sweden.'' This suggests that Vicuna-7B's inherent language modeling capabilities possibly override image reference interpretation.

Through this analysis, as explained in Section \ref{sec:problem_and_improvement}, we hypothesize two main challenges: (1) inadequate embeddings from the vision encoder and (2) a poor grasp of basic text comprehension tasks, crucial for complex document interpretation. These insights guide our strategic approach in developing \textsc{Elva}, addressing these challenges step by step.

\section{Experimental Details}
\label{sec:appendix_implementation}

\subsection{Software and Hardware Setup}
Our experiments are based on the official codebase\footnote{\url{https://github.com/haotian-liu/LLaVA}} of LLaVA~\citep{liu2023llava}. We utilize NVIDIA V100 32GB and A100 80GB GPUs for the computations. Ablation studies are conducted on V100 GPUs, whereas the final configuration models run on A100 GPUs. We do not observe any significant performance difference based on the type of GPU used. However, training on V100 GPUs is approximately 2 to 3 times slower per step compared to A100 GPUs. Although our codebase is based on LLaVA, to ensure better reproducibility, we will release the scripts used for training our models and any necessary code modifications as open-source.

\subsection{Datasets and Hyperparameters}\label{sec:hyperparameters}

\paragraph{Curated Dataset from LLaVA-1.5.}
Table~\ref{tab:llava_15_dataset} provides detailed quantities of the subsets within the dataset\footnote{\url{https://huggingface.co/datasets/liuhaotian/LLaVA-Instruct-150K/blob/main/llava_v1_5_mix665k.json}}.

\begin{table}[t]
\centering
\begin{adjustbox}{max width=0.4\linewidth}
\begin{tabular}{lc}
    \toprule
    \textbf{Dataset} & \textbf{\# Samples} \\
    \midrule
    LLaVA & 157,712 \\
    SG40k & 40,688 \\
    VQA-v2 & 82,783 \\
    GQA & 72,140 \\
    OKVQA & 8,998 \\
    OCRVQA & 80,000 \\
    A-OKVQA & 66,160 \\
    TextCaps & 21,953 \\
    RefCOCO & 48,447 \\
    VG & 86,417 \\
    \bottomrule
\end{tabular}
\end{adjustbox}
\caption{\textbf{Curated dataset from LLaVA-1.5.} Dataset proportions are shown. RefCOCO and VG are not used in the ablation studies.}\label{tab:llava_15_dataset}
\end{table}

\paragraph{Curated Dataset from \textsc{Elva}.}
Table \ref{tab:curated_elva_datasets} lists the datasets in \textsc{Elva}'s final configuration.
Meanwhile, for the alignment phase, we use the alignment datasets from LLaVA~\citep{liu2023llava} and LLaVAR~\citep{zhang2023llavar}, which consist of 558K and 422K samples respectively.

\begin{table}[t]
\centering
\begin{adjustbox}{max width=\linewidth}
\begin{tabular}{lccc}
    \toprule
    \textbf{Dataset} & \textbf{\# Samples} & \textbf{Sampling Ratio \%} \\
    \midrule
    LLaVA-1.5-Set (See Table~\ref{tab:llava_15_dataset}) & 665,298 & 46.49 \\
    Vision-Flan-Set~\cite{xu-etal-2024-vision} & 186,103 & 12.99 \\
    WikiArt~\cite{chen2023sharegpt4v} & 500 & 0.03 \\
    Celebrity~\cite{chen2023sharegpt4v} & 498 & 0.03 \\
    Landmark~\cite{chen2023sharegpt4v} & 500 & 0.03 \\
    Share-TextVQA~\cite{chen2023sharegpt4v} & 500 & 0.03 \\
    DocVQA~\cite{mathew2021docvqa} & 11,480 & 1.60 \\
    ChartQA~\cite{masry-etal-2022-chartqa} & 18,317 & 2.56 \\
    $^{\ast}$Cauldron-Set-AI2D & 2,434 & 0.17 \\
    $^{\ast}$Cauldron-Set-Chart2Text & 26,961 & 1.88 \\
    $^{\ast}$Cauldron-Set-Diagram-Image-to-Text & 300 & 0.02 \\
    $^{\ast}$Cauldron-Set-HITAB & 2,500 & 0.17 \\
    $^{\ast}$Cauldron-Set-IAM & 5,663 & 0.40 \\
    $^{\ast}$Cauldron-Set-RenderedText & 10,000 & 0.70 \\
    $^{\ast}$Cauldron-Set-Robut-SQA & 8,514 & 0.59 \\
    $^{\ast}$Cauldron-Set-Robut-WTQ & 38,246 & 2.67 \\
    $^{\ast}$Cauldron-Set-ScienceQA & 4,976 & 0.35 \\
    $^{\ast}$Cauldron-Set-Screen2words & 15,730 & 1.10 \\
    $^{\ast}$Cauldron-Set-STVQA & 17,247 & 1.20 \\
    $^{\ast}$Cauldron-Set-TabMWP & 22,722 & 1.59 \\
    $^{\ast}$Cauldron-Set-InfoVQA & 2,118 & 0.15 \\
    $^{\ast}$Cauldron-Set-TQA & 1,493 & 0.10 \\
    CORD-Instruct (Proposed in this work, \S\ref{sec:new_datasets}) & 680 & 0.05 \\
    VisualMRC~\cite{VisualMRC2021} & 7,959 & 0.56 \\
    LLaVAR-Inst~\cite{zhang2023llavar} & 19,732 & 1.38 \\
    DocReason~\cite{hu2024mplugdocowl} & 25,877 & 1.81 \\
    DocVQA-single$^{\dagger}$ & 44,815 & 9.38 \\
    ChartQA-single$^{\dagger}$ & 28,068 & 5.88 \\
    Layout-en-sampled$^{\ddagger}$~\cite{kim-etal-2023-visually} & 50,000 & 3.49 \\
    DVQA-sampled$^{\ddagger}$~\cite{kafle2018dvqa} & 10,000 & 0.70 \\
    MMC-Chart-sampled$^{\ddagger}$~\cite{liu-etal-2024-mmc} & 10,000 & 0.70 \\
    ScreenQA-sampled$^{\ddagger}$~\cite{hsiao2024screenqa} & 10,000 & 0.70 \\
    LRV-Chart-sampled$^{\ddagger}$~\cite{liu2023aligning} & 6,746 & 0.47 \\
    \bottomrule
\end{tabular}
\end{adjustbox}
\caption{\textbf{Overview of datasets used in the final data configuration.} All datasets are open-source and freely accessible. Datasets marked with $^{\ast}$ are subsets curated by \citet{laurencon2024matters}, with only selected portions adopted in this work. $^{\dagger}$ indicates that each question-answer pair is considered as a single sample. Datasets marked with $^{\ddagger}$ had a large volume of data, hence, only partial images were randomly sampled.}\label{tab:curated_elva_datasets}
\end{table}

\paragraph{Hyperparameters.}
Table~\ref{tab:hparam_align} and Table~\ref{tab:hparam_inst} provide the hyperparameters used during the alignment and instruction tuning stages, noting that smaller models benefit from larger learning rates. In the final model training configuration, we employ the data and sampling ratios outlined in Table~\ref{tab:curated_elva_datasets} and train the model for 11K steps. Calculating the exact number of unique images is complex due to overlap across datasets; however, we estimate using approximately 1M unique images. LLaVA-NeXT reported using 760K samples\footnote{\url{https://llava-vl.github.io/blog/2024-01-30-llava-next}}, and our use represents a modest increase. Furthermore, as we leverage multiple curated datasets with slightly different questions on the same images, we consider a synthetic epoch to consist of 1.4M examples. Thus, with a batch size of 128, we complete 11K steps (1.4M / 128). For ablation studies, we exclude datasets like VG, RefCOCO, and Vision-Flan to minimize costs, resulting in 9K training steps. Additionally, we find that the 0.2B model converges more slowly, so we extend its instruct tuning to twice the number of steps compared to the other models (1B to 13B). This increased number of steps is applied solely to the 0.2B scale.

\begin{table}[t]
\centering
\begin{adjustbox}{max width=\linewidth}
\begin{tabular}{l|cccccc}
    \toprule
    \textbf{Model Size} & \textbf{LR} & \textbf{Epsilon} & \textbf{Grad Clip Norm} & \textbf{Weight Decay} & \textbf{Warmup Ratio} \\
    \midrule
    0.2B & 1e-3 & 1e-6 & 0.5 & 0.0 & 0.03 \\
    1B & 1e-3 & 1e-6 & 0.5 & 0.0 & 0.03 \\
    3.8B & 1e-3 & 1e-6 & 0.5 & 0.0 & 0.03 \\
    7B & 1e-4 & 1e-6 & 0.5 & 0.0 & 0.03 \\
    13B & 1e-4 & 1e-6 & 0.5 & 0.0 & 0.03 \\
    \bottomrule
\end{tabular}
\end{adjustbox}
\caption{\textbf{Hyperparameters used during the alignment stage.}} \label{tab:hparam_align}
\end{table}

\begin{table}[t]
\centering
\begin{adjustbox}{max width=\linewidth}
\begin{tabular}{l|cccccc}
    \toprule
    \textbf{Model Size} & \textbf{LR} & \textbf{Epsilon} & \textbf{Grad Clip Norm} & \textbf{Weight Decay} & \textbf{Warmup Ratio} \\
    \midrule
    0.2B & 3e-4 & 1e-6 & 0.5 & 1e-3 & 0.03 \\
    1B & 2e-4 & 1e-6 & 0.5 & 1e-3 & 0.03 \\
    3.8B & 2e-4 & 1e-6 & 1.0 & 0.0 & 0.03 \\
    7B & 2e-5 & 1e-6 & 1.0 & 0.0 & 0.03 \\
    13B & 2e-5 & 1e-6 & 1.0 & 0.0 & 0.03 \\
    \bottomrule
\end{tabular}
\end{adjustbox}
\caption{\textbf{Hyperparameters used during the instruct tuning stage.} Larger learning rates were noted to be more effective for smaller models.}\label{tab:hparam_inst}
\end{table}

\subsection{Evaluation Details}

As described in Section \ref{sec:problem_and_improvement}, which outlines our strategic model development approach, we use eight benchmarks: {DocVQA} (\textbf{Doc})~\citep{mathew2021docvqa}, ChartQA (\textbf{Chart})~\citep{masry-etal-2022-chartqa}, {InfographicVQA} (\textbf{Info})~\citep{Mathew_2022_WACV}, {SEED-2-Plus} (\textbf{SD2P})~\citep{li2024seed2plus}, {SEED-IMG} (\textbf{SD-I})~\citep{li2023seed}, {MMStar} (\textbf{MMS})~\citep{chen2024we}, {ScienceQA-IMG} (\textbf{SciQA})~\citep{lu2022learn}, and {HallusionBench} (\textbf{Hall})~\citep{Guan_2024_CVPR}. Additionally, for the main experiments and analyses, we include several more benchmarks: \textbf{AI2D}~\citep{Kembhavi2016ADI}, MathVista-TestMini (\textbf{Math})~\citep{lu2024mathvista}, LLaVA-Bench (\textbf{LBen})~\citep{liu2023llava}, and the Parsing-Bench (\textbf{PBen}) proposed in this work. We conduct evaluations using \textit{VLMEvalKit}~\citep{2023opencompass} and the official code by \citet{liu2023llava}.
When evaluating LLaVA-Bench, we transition to using \texttt{gpt-4-0613} for judging, as the previously widely-used \texttt{gpt-4-0314} is deprecated.

DocVQA and InfographicVQA employ active leaderboards\footnote{\url{https://rrc.cvc.uab.es}}, which require JSON-formatted submissions for performance verification. This procedure, while thorough, can hinder rapid evaluations needed for iterative experimentation. To mitigate this, our ablation studies detailed in Section \ref{sec:remedies} utilize custom evaluation scripts for \textbf{Doc} and \textbf{Info}. This involves parsing ground truth data: for DocVQA, we extract information from CSV files provided by the leaderboard, while we use the validation set for InfographicVQA. Our custom evaluation scores show high correlation with the official leaderboard results, confirming their credibility. That is, for ablation studies, we adopt a simplified evaluation to effectively compare different architectures within the \textsc{Elva} framework. 

However, for the main results and analyses (e.g., Table \ref{tab:main_table}, \ref{tab:model_comparison}, \ref{tab:ablation_table_llavanext}, and \ref{tab:ablation_table_withocr}), which require comparisons with other models, we use test set performance metrics from the official leaderboard to ensure accurate, apple-to-apple comparisons.

\section{Details on \textsc{Elva}-Encoder Training}
\label{sec:appendix_vision_encoder}

\subsection{Dataset and Hyperparameters}
Our primary training focus is on text reading tasks, aimed at enhancing the vision encoder's text recognition capabilities. The training datasets include OCR-IDL~\citep{biten2022ocr} (837,922 samples), PDFA\footnote{\url{https://huggingface.co/datasets/pixparse/pdfa-eng-wds}} (1,048,569 samples), as well as the alignment sets from LLaVA~\citep{liu2023llava} and LLaVAR~\citep{zhang2023llavar}. We employ duplicate sampling, treating 3.5M samples as a synthetic epoch, completing one epoch as detailed in Table~\ref{tab:rencoder_12times}. This process is repeated to produce 12 distinct \textit{REncoder} variants. With the selected 1B model scale, each training session requires approximately 1.7 days using 8 V100 GPUs.

\begin{table}[t]
\centering
\begin{adjustbox}{max width=0.7\linewidth}
\begin{tabular}{ccc}
    \toprule
    \textbf{Batch Size} & \textbf{Learning Rate (LR)} & \textbf{Weight Decay} \\
    \midrule
    128 & 5e-5 & 1e-3 \\
    128 & 6e-5 & 1e-3 \\
    128 & 7e-5 & 1e-3 \\
    128 & 8e-5 & 1e-3 \\
    256 & 5e-5 & 1e-3 \\
    256 & 6e-5 & 1e-3 \\
    256 & 7e-5 & 1e-3 \\
    256 & 8e-5 & 1e-3 \\
    512 & 5e-5 & 0.0 \\
    512 & 6e-5 & 0.0 \\
    512 & 7e-5 & 0.0 \\
    512 & 8e-5 & 0.0 \\
    \bottomrule
\end{tabular}
\end{adjustbox}
\caption{\textbf{Configurations for 12 \textit{REncoder} trainings.} The table shows batch size, learning rate, and weight decay. All other hyperparameters are essentially identical to those in \S\ref{sec:hyperparameters}.}\label{tab:rencoder_12times}
\end{table}

\subsection{Details on Small VLM Usage}
As outlined in Section~\ref{sec:readclip}, each \textit{REncoder} variant undergoes training by unfreezing the vision encoder and fine-tuning it within a 1-billion-parameter model setting focused on text-centric datasets. We selected the 1B scale to balance computational demands and model performance. Preliminary experiments, detailed in Table~\ref{tab:small_vlm_experiments}, suggest that while a 0.2B \textsc{Elva}-Encoder configuration offers some benefits, the 1B variant yields noticeably better results. We attribute this improvement to the 1B model's superior text reading capabilities, enhancing learning outcomes during the \textsc{Elva}-Encoder process. Although coupling each REncoder with a larger model could further improve text recognition, it comes at the cost of increased computational resources and training time. For instance, a 3.8B model requires about 3.5 times longer to train than a 1B model, making it less feasible for many practitioners. Thus, the 1B parameter model was chosen to achieve significant enhancements while maintaining computational efficiency.

\begin{table}[t]
    \centering
    \begin{adjustbox}{max width=\linewidth}
    \begin{tabular}{l|ccc}
        \toprule
        \textbf{Target Size} & \textbf{No \textsc{Elva}-Encoder$^{\ast}$} & \textbf{With 0.2B Trained} & \textbf{With 1B Trained$^{\dagger}$} \\
        &  & \textbf{\textsc{Elva}-Encoder} & \textbf{\textsc{Elva}-Encoder}\\
        \midrule
        0.2B & 32.1 & \textbf{35.0} & 34.9 \\
        1B   & 43.0 & 44.4 & \textbf{45.5} \\
        3.8B & 47.4 & 49.4 & \textbf{50.2} \\
        \bottomrule
    \end{tabular}
    \end{adjustbox}
\caption{\textbf{Effect of \textsc{Elva}-Encoder integration.} This table illustrates the performance impact across various model sizes with and without \textsc{Elva}-Encoder integration, as well as the influence of VLM size used in \textsc{Elva}-Encoder training. Columns marked with $\ast$ correspond to results for C1 in Section 4.3, while those marked with $\dagger$ correspond to C5 in the same section.}
    \label{tab:small_vlm_experiments}
\end{table}

\begin{table}[t]
    \centering
    \begin{adjustbox}{max width=\linewidth}
    \begin{tabular}{l|ccc|c}
        \toprule
        \textbf{Ratio} & \textbf{Text-Centric} & \textbf{General} & \textbf{Overall} & \textbf{Desc.} \\
        \midrule
        0\% & 41.4 & 46.5 & 43.9 & \textit{REncoder} (Denoted as \textbf{C3} in Sec 4.3) \\
        5\% & 41.8 & 45.7 & 43.7 & \\
        7\% & 41.5 & 46.6 & 44.0 & \textbf{C7} in Sec 4.3 \\
        10\% & 41.3 & 46.5 & 43.9 & \\
        25\% & 41.2 & 46.5 & 43.9 & \\
        50\% & 41.2 & 49.1 & 45.1 & Avg (CLIP\&RE) (\textbf{C4} in Sec 4.3) \\
        100\% & 37.4 & 48.6 & 43.0 & CLIP-B-224-AnyRes (\textbf{C1} in Sec 4.3) \\
        \bottomrule
    \end{tabular}
    \end{adjustbox}
\caption{\textbf{Performance outcomes with varying weight averaging ratios.} This table illustrates the text-centric, general, and overall scores derived from different integration ratios of OpenAI CLIP weights and \textit{REncoder} contributions. Notable configurations such as \textbf{C1}, \textbf{C3}, \textbf{C7}, and \textbf{C4} are reported as detailed in Section 4.3.}
\label{tab:abl_merge_ratio}
\end{table}

\begin{table}[t]
    \centering
    \begin{adjustbox}{max width=\linewidth}
    \begin{tabular}{lcccccc}
        \toprule
        & \textbf{No merge} & \textbf{Avg (CLIP\&RE)} & \textbf{w/ 2REs} & \textbf{w/ 4REs} & \textbf{w/ 8REs} & \textbf{w/ 12REs} \\
        \midrule
        \textbf{Text} & \textbf{37.4} & \textbf{41.2} & \textbf{41.7} & \textbf{41.9} & \textbf{42.0} & \textbf{42.4} \\
        General & 48.6 & 49.1 & 47.6 & 45.1 & 46.3 & 48.6 \\
        Overall & 43.0 & 45.1 & 44.7 & 43.5 & 44.1 & 45.5 \\
        \bottomrule
    \end{tabular}
    \end{adjustbox}
\caption{\textbf{Effect of increasing the number of \textit{REncoders} on performance.} This table presents the text-centric, general, and overall scores resulting from utilizing different numbers of \textit{REncoder} integrations, detailing the observed trends in performance improvement across configurations.}
\label{tab:abl_num_rencoder}
\end{table}

\subsection{Details on Model Weight Averaging}
In this study, we propose an approach to develop an efficient and effective vision encoder by keeping the encoder's parameters learnable within a small VLM training framework. The resulting specialized vision encoder weights are merged with the original CLIP weights, known for general image understanding prowess. The idea of simply averaging multiple model weights has been shown to be effective in various studies~\citep{pmlr-v162-wortsman22a,jang2024modelstockneedjust}. Our approach closely relates to the \textit{Uniform Soup} method used as a baseline by \citet{pmlr-v162-wortsman22a}, where uniform merging is applied. Recently, more advanced weight merging techniques have been explored~\citep{pmlr-v162-wortsman22a,jang2024modelstockneedjust}. Given our framework's orthogonal nature to these methods, we expect our proposed practice to be complementary and potentially used in conjunction with these novel techniques.

In addition, we present supplementary experimental results conducted to design the experiment in Section \ref{sec:readclip}. These experiments are carried out using multiple 1B scale VLMs. First, we examine how different mixing ratios of OpenAI CLIP weights influence performance. It is observed that a balanced improvement in performance occurs across various ratios (see Table~\ref{tab:abl_merge_ratio}). Subsequently, we increase the number of \textit{REncoder} instances used in merging, with results presented in Table~\ref{tab:abl_num_rencoder}. As shown, performance generally improves as more \textit{REncoder} instances are included, with our decision ultimately favoring the use of 12 encoders as a promising configuration. Importantly, this does not imply that 12 is an optimal or necessary quantity, as the use of even a single trained encoder followed by weight averaging results in appealing performance. With weight merging algorithms continuing to advance~\citep{pmlr-v162-wortsman22a}, we believe achieving higher performance with lower training costs is feasible and a potential direction for future work. Given our framework's orthogonal nature to these methods, we expect our proposed practice to be complementary and potentially used in conjunction with these novel techniques~\citep{pmlr-v162-wortsman22a,jang2024modelstockneedjust}.

\begin{table*}[!t]
\centering
\begin{adjustbox}{max width=0.83\linewidth}
\begin{tabular}{l}
\toprule
\textbf{Prompt} \\
\midrule
Carefully decipher the text in this image. Provide the text in the image only. \\
Investigate the image for any text. Provide the text in the image only. \\
Examine the image for any letters or words. Provide the text in the image only. \\
Identify all written characters present in the image. Provide the text in the image only. \\
Do a careful reading of the image and transcribe all text. Provide the text in the image only. \\
Inspect the image and write down all readable characters. Provide the text in the image only. \\
Translate the image content into written text. Provide the text in the image only. \\
Review the image and offer a transcription of the text. Provide the text in the image only. \\
Look over the image and jot down all visible text. Provide the text in the image only. \\
Scrutinize the image for any discernible words or letters. Provide the text in the image only. \\
Study the image and document all characters found within. Provide the text in the image only. \\
Assess the image and record any visible words or letters. Provide the text in the image only. \\
Kindly extract any text or characters from this image. Provide the text in the image only. \\
Evaluate the image and isolate all written content. Provide the text in the image only. \\
Assimilate all readable characters within the image. Provide the text in the image only. \\
Decode any legible text visible in the image. Provide the text in the image only. \\
Grasp all written elements within the image. Provide the text in the image only. \\
Conduct a thorough examination of the image and capture all text. Provide the text in the image only. \\
Peel all readable characters out from the image. Provide the text in the image only. \\
Interpret all characters from the image. Provide the text in the image only. \\
Understand and transcribe any textual content from the image. Provide the text in the image only. \\
Dissect the image and recover any written characters. Provide the text in the image only. \\
Deduce and present all visible text from the image. Provide the text in the image only. \\
Inspect the image and translate any visible characters into text. Provide the text in the image only. \\
\bottomrule
\end{tabular}
\end{adjustbox}
\caption{\textbf{Prompt examples for text reading tasks used in RR-Prompt.}}
\label{tab:rr_prompt_examples}
\end{table*}

\begin{table}[!t]
\centering
\begin{adjustbox}{max width=\linewidth}
\begin{tabular}{l}
\toprule
\textbf{A data sample from our train set} \\ 
\midrule
\textbf{User:} Deduce and present all visible text from the image. \\ 
\cmidrule(lr){1-1}
\textbf{Assistant:} 13 PAKET WEEKEND 1 55.000 715.000 6 PAKET\\
WEEKEND 2 (...) 6 Qty: 23 Tunai 1.250.000 Kembali 29.500\\ 
\cmidrule(lr){1-1}
\textbf{User:} Can you provide the menu items extracted from the docu-\\
ment image in Markdown format? \\ 
\cmidrule(lr){1-1}
\textbf{Assistant:} Sure, here are the menu items: \\
- PAKET WEEKEND 1: \$55.000 x 13 = \$715.000 \\
- PAKET WEEKEND 2: \$80.000 x 6 = \$480.000 \\
- ES JERUK: \$13.000 \\
- ES TEH TAWAR: \$3.000 \\
- KOPI HITAM: \$8.500 \\
- ES BATU: \$1.000 \\ 
\bottomrule
\end{tabular}
\end{adjustbox}
\caption{\textbf{An example data sample from our train set with RR-Prompt applied.}}
\label{tab:train_set_sample}
\end{table}

\section{Details on RR-Prompt}
\label{sec:appendix_rr_prompts}

In our study, we employ the RR-Prompt strategy to enhance the text understanding capability of the \textsc{Elva} model. This strategy involves inserting an initial QA turn for text-rich images requiring reasoning, prompting the model to first identify the text within the image. This approach ensures that the model reads the text before engaging in complex reasoning within a dialogue scenario. We apply the RR-Prompt selectively to specific text-rich datasets from our curated dataset, as shown in Table~\ref{tab:curated_elva_datasets}, ensuring that not all text-containing datasets are affected, to avoid potential mismatches.

Even within applicable subsets, RR-Prompt is not uniformly applied to all samples; samples with too little or too much text are excluded, and 20\% of samples are randomly skipped to help the model balance between reading all text when prompted and performing direct reasoning during inference. This selective application enables robust operation out-of-the-box without notable mismatches. To prevent over-specialization to a single OCR engine, we generate annotations using a combination of MS OCR\footnote{\url{https://docs.microsoft.com/en-us/azure/cognitive-services/computer-vision/overview-ocr}} and CLOVA OCR\footnote{\url{https://clova.ai/ocr/en}}. Processing text-heavy samples, such as those in DocVQA, takes approximately 4 seconds per call using the CLOVA OCR API. However, we anticipate optimizing this cost in future iterations.

As shown in Table \ref{tab:rr_prompt_examples}, the RR-Prompt incorporates an initial QA turn that instructs the model to read the text using straightforward commands. Furthermore, Table \ref{tab:train_set_sample} presents a sample where the RR-Prompt is applied. Despite its simplicity, this approach significantly enhances the training outcomes.

\begin{table}[t!]
\centering
\begin{tcolorbox}[fontupper=\small, title=\small CORD-Instruct Generation Prompt]
Create a synthetic user query requesting information extraction from a given document image. The extracted information should be provided in various formats such as JSON, XML, or Markdown based on the user's request. Users may ask for specific parts of the information, like menu items or payment amounts. Your task is to generate both a user query and the corresponding response from the information extraction system. Ensure the queries and responses vary in detail and format. Sometimes include concise responses, particularly when indicated with the word "concisely."

The provided JSON is to guide your response creation - do not display or mention it in the user queries. Specifically, do not create queries that ask for information extraction from a provided JSON (e.g., "Can you extract the information from the provided JSON" is not allowed). Additionally, you do not need to strictly follow the tag names in the provided JSON while creating your responses (e.g., "nm" can be "name" or "cnt" can be "count" ). Return results strictly in the format shown below:

\begin{verbatim}
Query: I need the payment amount 
from the document in this image 
in JSON format, answer concisely.
Answer: <<<json
{
  "payment_amount": "$123.45"
}
>>>
\end{verbatim}

\begin{verbatim}
Query: Please parse the input document
and provide the menu details in XML.
Answer: Certainly, here is the menu 
information in XML format:...
\end{verbatim}

\begin{verbatim}
Query: Can you provide the extracted
customer information from the document
image in Markdown?
Answer: Sure, here it is:
<<<
# Customer Information
- Name: John Doe
- Email: john.doe@example.com
>>>
\end{verbatim}

Remember, the goal is to include appropriate formatting such as JSON, XML, or Markdown in your responses to correspond with the user's query.
\end{tcolorbox}
\caption{\textbf{CORD-Instruct data generation prompt.}}
\label{tab:prompt}
\end{table}

\section{Details on the Construction of CORD-Instruct and Parsing-Bench}
\label{sec:appendix_new_datasets}

\subsection{CORD-Instruct}

The Consolidated Receipt Dataset (CORD), introduced by \citet{park2019cord}, is crafted for post-OCR parsing tasks, featuring Indonesian receipt images with structured data in JSON format. Our goal is to adapt this dataset to scenarios where models must generate user-requested structured information in formats like JSON, XML, or Markdown. To achieve this, we construct CORD-Instruct using the OpenAI GPT-3.5 API\footnote{Specifically, \texttt{gpt-3.5-turbo-0125}.}. A detailed prompt, presented in Table \ref{tab:prompt}, guides the API in generating synthetic user queries and corresponding responses based on the provided structured information.

\begin{figure}[t!]
    \centering
    \includegraphics[width=\linewidth]{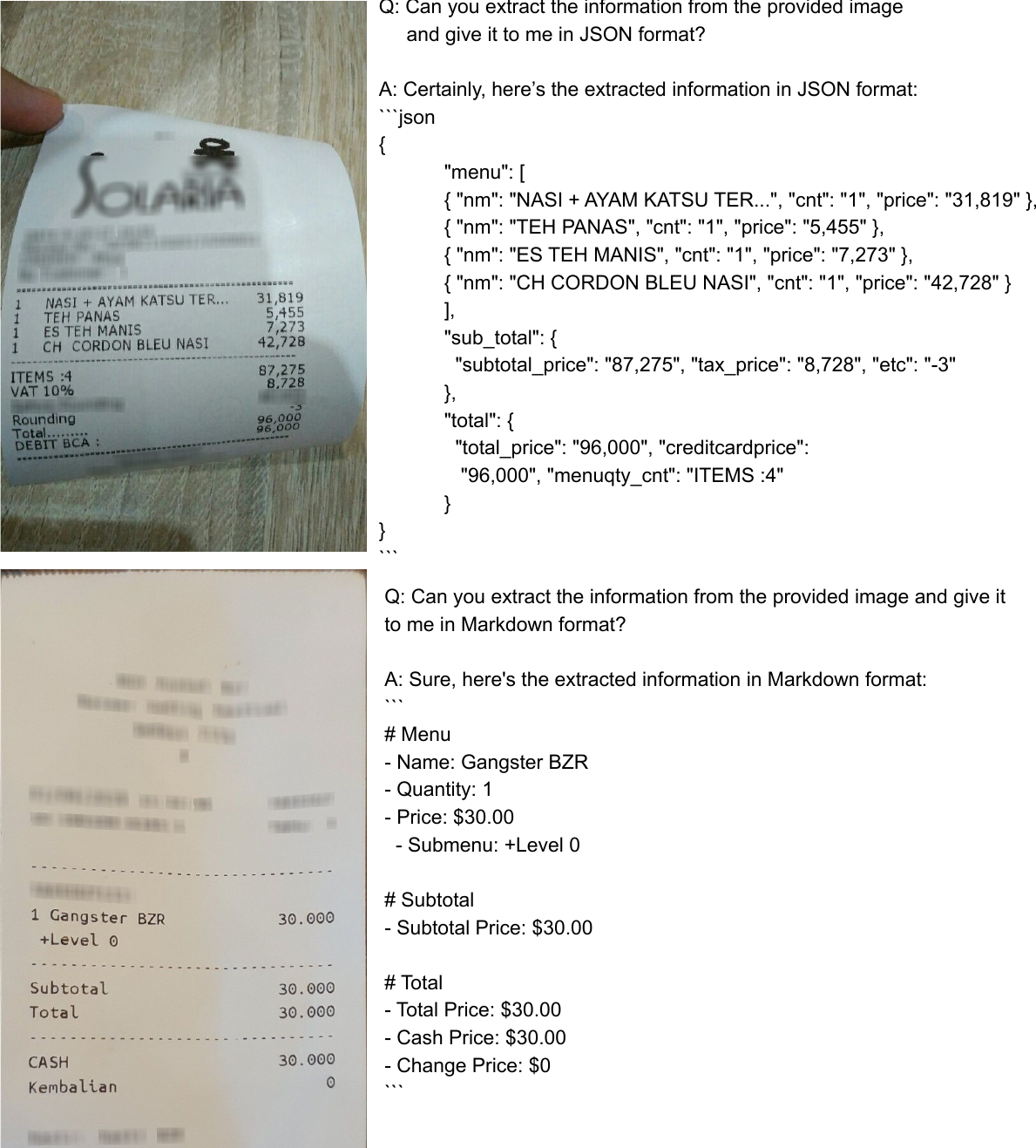}
    \caption{\textbf{The generated CORD-Instruct examples.}}
    \label{fig:cord}
\end{figure}

After generating initial instructions and API responses, we manually filter out any unsuccessful samples. This process involves verifying that samples reflect real-world scenarios, removing ambiguous instructions, and correcting errors in JSON, XML, or Markdown transcriptions. Examples of the generated CORD-Instruct data are shown in Figure~\ref{fig:cord}.

\begin{table}[!t]
\centering
\begin{tcolorbox}[fontupper=\small, title=\small Parsing-Bench BID Context Generation Prompt]
Using the provided Brazilian Identity Document image, please compose a comprehensive and detailed caption that encapsulates all the elements depicted in the image. Ensure precision in extracting any text present, maintaining case sensitivity and retaining the exact original form. Begin with a well-written caption in natural language, detailing the image's content, layout, and nuances. Conclude with a well-structured XML format that meticulously documents the extracted information, preserving the image's original layout and details.
\end{tcolorbox}
\begin{tcolorbox}[fontupper=\small, title=\small Parsing-Bench SROIE Context Generation Prompt]
Using the provided scanned receipt image, please compose a comprehensive and detailed caption that encapsulates all the elements depicted in the image. Ensure precision in extracting any text present, maintaining case sensitivity and retaining the exact original form. Begin with a well-written caption in natural language, detailing the image's content, layout, and nuances. Conclude with a well-structured XML format that meticulously documents the extracted information, preserving the image's original layout and details.
\end{tcolorbox}
\begin{tcolorbox}[fontupper=\small, title=\small Parsing-Bench Evaluation Rubric]
We would like to request your feedback on the performance of two AI assistants in response to the user's question displayed above. The user asks the question related to the document image. For your reference, the visual content in the document image is described with a caption, and a corresponding XML file summarizes the information. Please note that the information provided in the caption and XML file may not be completely perfect. Please rate the helpfulness, relevance, accuracy, and level of detail of their responses. Each assistant receives an overall score on a scale of 1 to 10, where a higher score indicates better overall performance. Please first output a single line containing only two values indicating the scores for Assistant 1 and 2, respectively. The two scores should be separated by a space. In the subsequent line, please provide a comprehensive explanation of your evaluation, avoiding any potential bias and ensuring that the order in which the responses were presented does not affect your judgment. Keep in mind that the XML file is a summarized version of the full information and may not be directly related to the user's question. Providing unrelated information will not be particularly beneficial. Responses should directly address the user's query in a clear and useful manner to achieve a higher score. Focus especially on whether the responses would be convenient and useful for the user. Minor typographical errors should not heavily impact the scoring. When parsing data formats like XML or JSON, if the user hasn't specified a particular format, minor differences in tag or key names are acceptable as long as the overall meaning is preserved.
\end{tcolorbox}
\caption{\textbf{Parsing-Bench context generation prompts and LLM-as-a-Judge evaluation rules.}}
\label{tab:prompt_parse1}
\end{table}

\begin{figure*}[!t]
    \centering
    \includegraphics[width=\linewidth]{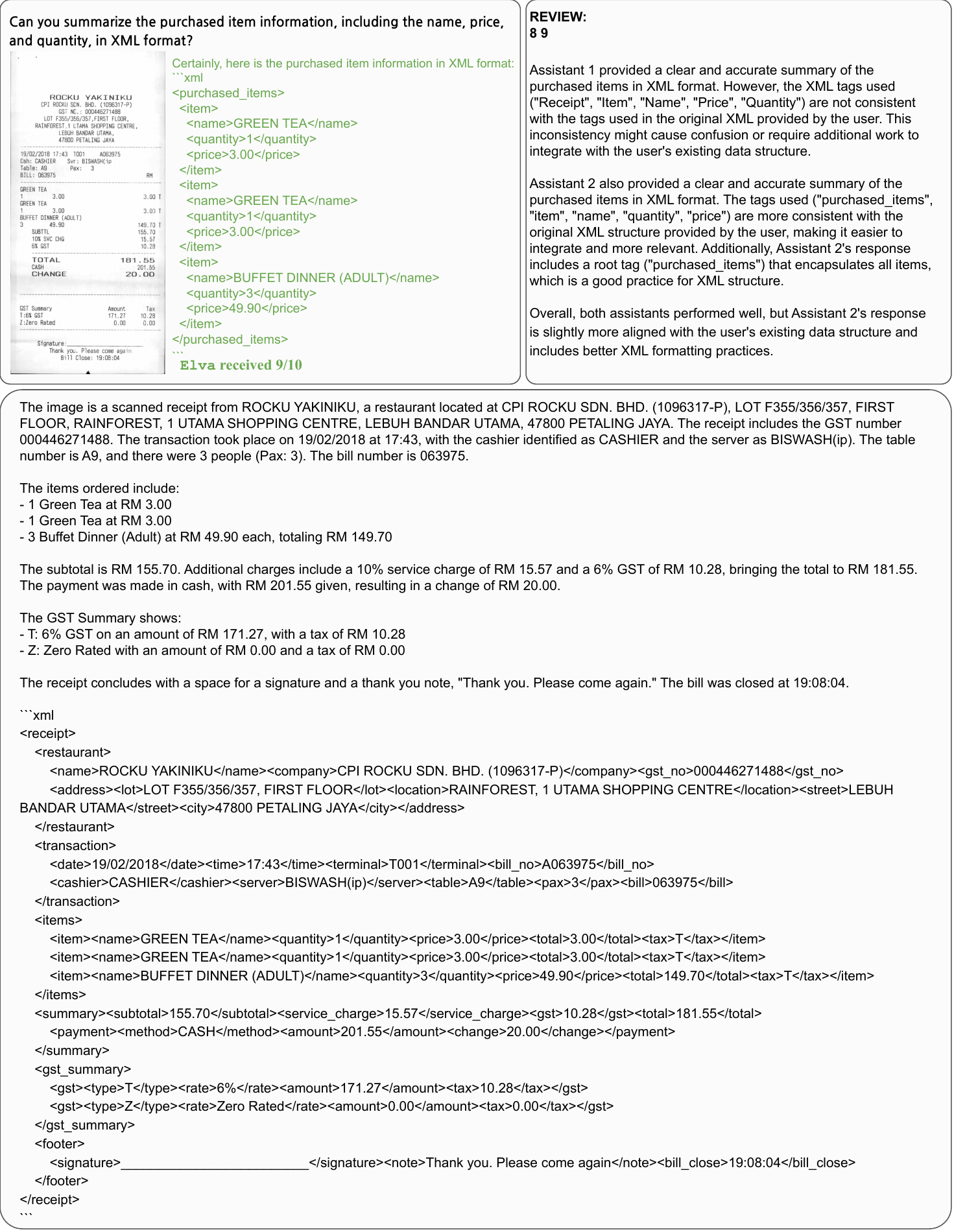}
\caption{\textbf{Overview of Parsing-Bench with an example.} The top left shows the question and model predictions. The top right contains the evaluation review. The bottom section, discussed in Section~\ref{sec:parse_bench_detail}, shows context extracted from the image using prompts. Evaluation compares two model predictions input into a high-performance LLM judge model. In this example, \textsc{Elva} is Assistant 2, highlighting its comparative performance.}
    \label{fig:parse_bench_detail}
\end{figure*}

\subsection{Parsing-Bench}\label{sec:parse_bench_detail}

Parsing-Bench is a dataset designed to fulfill the practical needs of visual document assistants. Inspired by LLaVA-Bench~\citep{liu2023llava}, this task requires the model to accurately interpret and analyze input document images to generate the desired structured output. Many industries have expressed a need to extract specific information from document images and convert it into formats like JSON or XML~\citep{kim2022donut}. Parsing-Bench uses Brazilian Identity Documents (BID)~\citep{sibgrapi_estendido} and SROIE~\citep{8977955} as image sources. Figure~\ref{fig:parse_bench_detail} visually depicts the overall evaluation process for better comprehension of our method.

To create the benchmark, we first extract context information from images, with prompts detailed in Table~\ref{tab:prompt_parse1}. An example is shown in Figure~\ref{fig:parse_bench_detail}. This extracted context is utilized during the LLM-as-a-Judge~\citep{zheng2023judging} evaluation process.

We prepare reference answers using the OpenAI GPT-4o API model\footnote{Specifically, \texttt{gpt-4o-2024-05-13}.} to serve as benchmarks for evaluation. Evaluation rules, or the rubric, are detailed in Table~\ref{tab:prompt_parse1}. Finally, the judge model, utilizing the OpenAI GPT-4o API, evaluates the target model outputs by comparing them against these reference answers, using the rubric and context to determine performance scores.

Unlike traditional benchmarks relying on rigid rule-based evaluations, Parsing-Bench offers more adaptability for assessing MLLMs. Parsing-Bench includes 30 examples that test models' comprehension and reasoning from document images. We believe that future work can expand Parsing-Bench by increasing the number of examples and encompassing a wider variety of documents and scenarios, enhancing its robustness and applicability. In line with our commitment to open research, we will make these datasets publicly available at \url{https://github.com/naver-ai/elva}.

\end{document}

%% file: latex/ablation_table_readclip.tex
\begin{table}[t!]
    \centering
    \begin{adjustbox}{max width=\linewidth}
    \begin{tabular}{lccc}
        \toprule
        \textbf{Vision Encoder Configuration} & \textbf{Text-Centric} & \textbf{General} & \textbf{Overall} \\
        \midrule
        \textbf{C1.} CLIP-B-224-AnyRes (CLIP) & 40.3 & 54.5 & 47.4 \\
        \textbf{C2.} Unfreeze CLIP        & 34.1 & 47.6 & 40.9 \\
        \textbf{C3.} \textit{REncoder} (RE)     & 45.2 & 52.2 & 48.7 \\
        \textbf{C4.} Avg (CLIP\& RE) & 45.6 & 54.4 & \underline{50.0} \\
        \rowcolor{maroon!15} \textbf{C5.} \textbf{\textsc{Elva}-encoder} (Avg (CLIP\& 12 REs)) & \underline{45.7} & \underline{54.7} & \textbf{50.2} \\
        \midrule
        \midrule
        \multicolumn{2}{l}{\textit{Supplementary ablations}} \\
        \textbf{C6.} CLIP-L-336 (\textit{LLaVA-1.5 on our data})   & 37.5 & \textbf{58.6} & 48.1 \\
        \textbf{C7.} CLIP (7\%) + \textit{REncoder} (93\%)  & \textbf{45.9} & 53.5 & {49.7} \\
        \bottomrule
    \end{tabular}
    \end{adjustbox}
    \caption{\textbf{Ablation study results for different vision encoder configurations.} Average scores for text-centric tasks (DocVQA, ChartQA, InfoVQA, and SEED-2-Plus), and general image tasks (SEED, MMStar, ScienceQA, and HallusionBench) are reported. These results are obtained with Phi-3 (3.8B). The overall scores for other scales (from 1B to 13B) are shown in Figure~\ref{fig:ve_ablation_larger_scales}.}
    \label{tab:ve_ablations_updated}
\end{table}

%% file: latex/abltion_table_rr.tex
\begin{table}[t!]
    \centering
    \begin{adjustbox}{max width=\linewidth}
    \begin{tabular}{lcccc}
        \toprule
         &1B & 3.8B & 7B & 13B \\
         & Text/Gen/All & Text/Gen/All & Text/Gen/All & Text/Gen/All \\
        \midrule
        \textbf{R1.} & 41.6/48.7/45.2 & 43.7/53.9/48.8 & 47.5/56.0/51.8 & 48.5/56.5/52.5 \\
        \rowcolor{maroon!15} \textbf{R2.} & \textbf{42.4}/48.6/\textbf{45.5} & \textbf{45.7}/\textbf{54.7}/\textbf{50.2} & \textbf{49.2}/\textbf{56.6}/\textbf{52.9} & \textbf{50.4}/57.4/\textbf{53.9} \\
        \midrule
        \midrule
        \multicolumn{5}{l}{\textit{Supplementary ablations}} \\
        \textbf{R3.} & 41.2/46.5/43.9 & 45.4/53.7/49.6 & 47.0/55.3/51.2 & 50.8/56.1/53.5 \\
        \textbf{R4.} & 42.0/\textbf{48.8}/45.4 & 44.1/54.2/49.1 & 48.2/56.0/52.1 & 49.6/\textbf{57.6}/53.6 \\
        \bottomrule
    \end{tabular}
    \end{adjustbox}
    \caption{\textbf{Comparison of different model sizes and RR-Prompt variants.} \textbf{R1} represents standard models trained without additional text reading prompt. \textbf{R2} employs explicit initial text reading steps for text-rich taks. \textbf{R3} carries out text reading at the end, while \textbf{R4} provides OCR results just as context without explicit supervision.}
    \label{tab:rr_ablations}
\end{table}

%% file: latex/main_table.tex
\begin{table*}[!th]
\begin{adjustbox}{width=\linewidth}
\setlength\tabcolsep{3pt} 
\renewcommand{\arraystretch}{1.2} 
\begin{tabular}{l|rrrrr|ccccc|cccccccc}
\toprule
\multirow{2}{*}{\textbf{Model}} & \multicolumn{2}{c}{\textbf{\# Param}} & \multirow{2}{*}{\textbf{\#tok}} & \multirow{2}{*}{\textbf{s/img}} & \multirow{2}{*}{\textbf{vram}} & \multicolumn{5}{c}{\textit{Text-Centric Benchmarks}} & \multicolumn{7}{c}{\textit{General Multimodal Benchmarks}} \\
           & {Vision} & {LM} &  &  &  & \textbf{Doc} & \textbf{Chart} & \textbf{Info} & \textbf{SD2P} & \textbf{PBen} & \textbf{SD-I} & \textbf{MMS} & \textbf{SciQA} & \textbf{Hall} & \textbf{AI2D} & \textbf{Math} & \textbf{LBen} \\
\midrule
LLaVA-v1-13B             & 300M & 13B & 576 & 1.43 & 26.9 & 9.8 & 7.0 & 19.9 & 39.5 & 14.0 & 51.2 & 32.9 & 62.4 & 43.0 & 43.9 & 25.9 & 69.9 \\
LLaVA-1.5-7B             & 300M & 7B & 576 & 0.46 & 14.7 & 22.8 & 17.8 & 22.4 & 41.2 & 17.9 & 65.9 & 33.1 & 69.2 & 48.5 & 55.6 &25.6 & 59.6 \\
LLaVA-1.5-13B             & 300M & 13B & 576 & 0.66 & 27.0 & 24.5 & 18.5 & 24.9 & 44.4 & 19.6 & 68.2 & 34.1 & {72.3} & 45.7 & 60.7 & 27.7 & 66.1 \\
LLaVA-NeXT-7B            & 300M & 7B & 1728-2880 & 1.01 & 20.0 & 68.3 & 51.9 & 31.6 & 51.7 & 49.6 & 69.8 & 38.2 & 69.0 & 44.8 & 66.8 & 31.8 & \textbf{72.3} \\
LLaVA-NeXT-13B           & 300M & 13B & 1728-2880 & 1.78 & 40.1 & \underline{69.8} & 59.0 & \textbf{34.9} & \textbf{55.6} & \underline{57.3} & \textbf{71.5} & \underline{41.2} & 73.4 & 46.7 & \textbf{71.7} & 34.1 & \textbf{72.3} \\
\rowcolor{maroon!15} \textbf{\textsc{Elva}-0.2B (ours)}        & 88M & 0.2B & 98-637 & 0.24 & 1.4 & 44.7 & 50.3 & 14.8 & 31.4 & 12.3 & 37.8 & 31.5 & 39.0 & 48.1 & 31.0 & 27.0 & 28.4 \\
\rowcolor{maroon!15} \textbf{\textsc{Elva}-1B (ours)}           & 88M & 1B & 98-637 & 0.41 & 3.3 & 62.6 & 57.7 & 23.7 & 36.8 & 27.3 & 52.3 & 32.6 & 63.3 & 50.4 & 46.9 & 31.7 & 36.0 \\

\rowcolor{maroon!15} \textbf{\textsc{Elva}-3.8B (ours)}         & 88M & 3.8B & 98-637 & 0.45 & 9.1 & 66.1 & \underline{61.9} & 24.2 & 44.7 & 31.0 & 61.3 & 36.9 & 74.2 & 52.7 & 63.0 & 35.6 & 45.3 \\ %

\rowcolor{maroon!15} \textbf{\textsc{Elva}-7B (ours)}           & 88M & 7B & 98-637 & 0.54 & 14.5 & 69.1 & 61.8 & 30.7 & 47.7 & 45.0 & 62.6 & 35.4 & 74.7 & \textbf{56.8} & 66.2 &  \underline{36.6} & 50.7 \\
\rowcolor{maroon!15} \textbf{\textsc{Elva}-13B (ours)}          & 88M & 13B & 98-637 & 0.72 & 27.0 & \textbf{71.7} & \textbf{65.2} & \underline{34.6} & \underline{52.6} & \textbf{59.2} & 65.3 & 37.9 & \underline{77.7} & \textbf{56.8} & \underline{69.3} & \textbf{38.1} & 51.0 \\
\midrule
\midrule
\multicolumn{17}{l}{\textit{Supplementary baselines}} \\
Qwen-VL-7B                & 1882M & 7B & 224 & 0.50 & 19.2 & 65.1 & 60.2 & -- & 41.0 & -- & 56.5 & 33.9 & 60.6 & 37.4 & 57.2 & 15.5 & 12.9 \\
Qwen-VL-7B-Chat           & 1882M & 7B & 224 & 0.56 & 19.2 & 62.6 & 49.3 & -- & 46.9 & -- & 62.9 & 34.0 & 64.0 & 40.8 & 59.7 & 34.9 & 67.7 \\
PaliGemma-3B          & 428M & 3B & 1024 & 0.98 & 10.3 & -- & 33.8 & -- & 49.8 & -- & \underline{70.0} & \textbf{48.6} & \textbf{94.3} & 53.0 & \underline{69.3} & 28.7 & 36.9 \\
\bottomrule
\end{tabular}
\end{adjustbox}
\caption{\textbf{Performance comparison across different models and benchmarks.} This table summarizes model sizes (Vision and LM), token counts (\#tok), latency (s/img), and memory cost (vram). The performance metrics across various benchmarks are presented, showcasing each model's strengths and weaknesses in different challenges.}
    \label{tab:main_table}
\end{table*}

%% file: latex/ablation_table_llavanext.tex
\begin{table}[t!]
  \centering
  \begin{adjustbox}{width=0.95\linewidth}
    \begin{tabular}{l|rr|cccc}
      \toprule
      \textbf{Method} & \textbf{s/img} & \textbf{vram} & \textbf{Doc} & \textbf{Chart} & \textbf{Info} & \textbf{SD2P} \\
      \midrule
      LLaVA-NeXT-7B & 1.01 & 20.0 & \underline{68.3} & \underline{51.9} & \textbf{31.6} & \textbf{51.7} \\
      \quad -- w/ max. 1728 tokens & \underline{0.70} & \underline{17.1} & 51.7 & 48.0 & 27.9 & 44.9 \\
      \rowcolor{maroon!15} \textbf{\textsc{Elva}-7B (ours)} & \textbf{0.54} & \textbf{14.5} & \textbf{69.1} & \textbf{61.8} & \underline{30.7} & \underline{47.7} \\
      \midrule
      \midrule
      LLaVA-NeXT-13B & 1.78 & 40.1 & \underline{69.8} & \underline{59.0} & \textbf{34.9} & \textbf{55.6} \\
      \quad -- w/ max. 1728 tokens & \underline{1.11} & \underline{30.4} & 53.9 & 52.3 & 30.9 & 49.2 \\
      \rowcolor{maroon!15} \textbf{\textsc{Elva}-13B (ours)} & \textbf{0.72} & \textbf{27.0} & \textbf{71.7} & \textbf{65.2} & \underline{34.6} & \underline{52.6} \\
      \bottomrule
    \end{tabular}
  \end{adjustbox}
  \caption{\textbf{Ablations on reduced vision token counts.} Given time and memory costs, \textsc{Elva} shows benefits.}
  \label{tab:ablation_table_llavanext}
\end{table}

%% file: latex/ablation_table_withocr.tex
\begin{table}[t!]
  \centering
  \begin{adjustbox}{width=\linewidth}
    \begin{tabular}{l|cccc}
      \toprule
      \textbf{Method} & \textbf{Doc} & \textbf{Chart} & \textbf{Info} & \textbf{SD2P} \\
      LLaVA-NeXT-7B & 74.5 (↑6.2) & 53.7 (↑1.8) & 35.5 (↑3.9) & 55.3 (↑3.6) \\
      \rowcolor{maroon!15} \textbf{\textsc{Elva}-7B (ours)} & \textbf{77.8} (↑8.7) & \textbf{64.0} (↑2.2) & \textbf{39.5} (↑8.8) & \textbf{55.7} (↑8.0) \\
      \midrule
      \midrule
      LLaVA-NeXT-13B & 76.5 (↑6.7) & 62.5 (↑3.5) & 40.4 (↑5.5) & 58.9 (↑3.3) \\
      \rowcolor{maroon!15} \textbf{\textsc{Elva}-13B (ours)} & \textbf{81.1} (↑9.4) & \textbf{67.5} (↑2.3) & \textbf{44.8} (↑10.2) & \textbf{60.6} (↑8.0) \\
      \bottomrule
    \end{tabular}
  \end{adjustbox}
\caption{\textbf{Performance gains with OCR integration.} \textsc{Elva} excels in both OCR-free and OCR-based modes.}
  \label{tab:ablation_table_withocr}
\end{table}